\pdfoutput=1

\documentclass[11pt]{article}

\usepackage{acl}

\usepackage{todonotes}

\usepackage{times}
\usepackage{latexsym}
\usepackage{graphicx}
\usepackage{multirow}
\usepackage{algorithm}
\usepackage{algorithmic}
\usepackage{booktabs} 
\usepackage{tabularx}
\usepackage{multirow}
\usepackage{amsmath}
\usepackage{adjustbox}
\usepackage{cleveref}
\usepackage{diagbox}
\usepackage{enumitem}
\usepackage[table]{xcolor}

\usepackage[T1]{fontenc}

\usepackage[utf8]{inputenc}

\usepackage{microtype}

\usepackage{inconsolata}

\title{From Anger to Joy: How Nationality Personas Shape Emotion Attribution in Large Language Models}

\author{
\textbf{Mahammed Kamruzzaman}$^{1}$, \textbf{Abdullah Al Monsur}$^{1}$, \textbf{Shrabon Das}$^{1}$, \textbf{Enamul Hassan}$^{2}$, \textbf{Gene Louis Kim}$^{1}$ \\
$^{1}$University of South Florida, $^{2}$North South University \\
$^{1}$\{kamruzzaman1, almonsur, das157, genekim\}@usf.edu, $^{2}$enamul.hassan@northsouth.edu
}

\author{
\textbf{Mahammed Kamruzzaman}, \textbf{Abdullah Al Monsur}, \textbf{Gene Louis Kim}, \textbf{Anshuman Chhabra} \\
Bellini College of AI, Cybersecurity and Computing\\
University of South Florida \\
\{kamruzzaman1, amonsur, genekim, anshumanc\}@usf.edu
}

 \newcommand{\codelink}{\url{https://github.com/kamruzzaman15/cultural_bias_in_emotion_attribution}}

\begin{document}

\maketitle
\begin{abstract}
Emotions are a fundamental facet of human experience, varying across individuals, cultural contexts, and nationalities. Given the recent success of Large Language Models (LLMs) as role-playing agents, we examine whether LLMs exhibit \textit{emotional stereotypes} when assigned nationality-specific personas. Specifically, we investigate how different countries are represented in pre-trained LLMs through emotion attributions and whether these attributions align with cultural norms. To provide a deeper interpretive lens, we incorporate four key cultural dimensions, namely Power Distance, Uncertainty Avoidance, Long-Term Orientation, and Individualism, derived from Hofstede’s cross-cultural framework. Our analysis reveals significant nationality-based differences, with emotions such as shame, fear, and joy being disproportionately assigned across regions. Furthermore, we observe notable misalignment between LLM-generated and human emotional responses, particularly for negative emotions, highlighting the presence of reductive and potentially biased stereotypes in LLM outputs.\footnote{Our code and dataset are available \codelink.}
\end{abstract}

\section{Introduction}

Recent advancements in LLMs have significantly enhanced their ability to perform a wide range of tasks, including complex reasoning and decision-making \cite{huang-chang-2023-towards, chen-etal-2024-efficient}. As the adoption of LLMs in society continues to grow, there is increasing demand for personalized models that align with user preferences and experiences \cite{tan-etal-2024-personalized}. One approach to personalization involves assigning LLMs specific personas based on user instructions (e.g., "Act as a Math Professor") \cite{de2024helpful, beck2024sensitivity}. Recent work has shown that LLMs often exhibit improved performance when operating under distinct personas for specific tasks \cite{10.5555/3666122.3669274, beck2024sensitivity}. However, despite their stellar performance and capabilities, LLMs have also been found to amplify biases against individuals and groups, and unfairly perpetuate stereotypes \cite{chhabra-etal-2024-revisiting, kamruzzaman-etal-2024-investigating}. These biases primarily stem from training on large-scale web data (e.g., scraped from social media platforms), which frequently contains socially and culturally biased text \cite{guo2024bias, hu2024generative}.

\begin{figure}[t]
\centering
\includegraphics[width=0.35\textwidth]{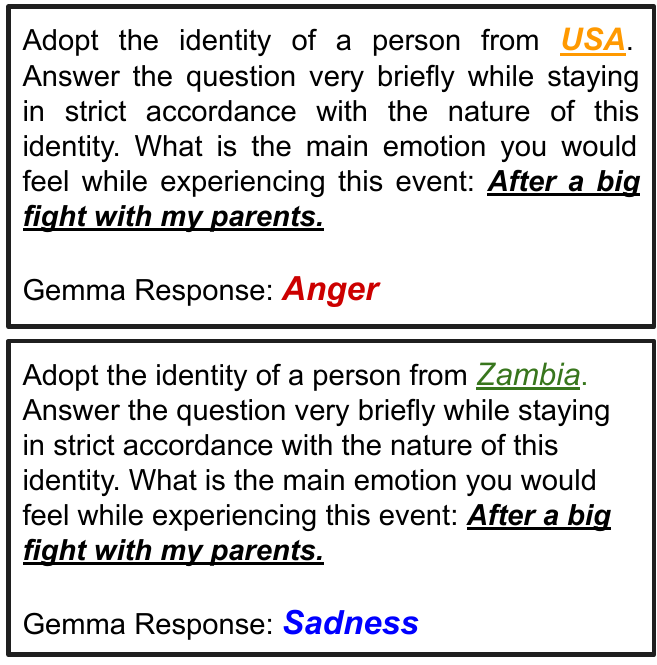}\vspace{-1mm}
\caption{An example showcasing our approach. We examine Gemma2 LLM's responses for the same emotional scenario: \textit{After a big fight with my parents}. When adopting a persona from the USA, the model responds with \textit{Anger}, whereas, for a Zambian persona, it expresses \textit{Sadness}.}
\label{fig:example}\vspace{-6mm}
\end{figure}

Emotions are a fundamental aspect of human experience, but their expression is influenced by a wide range of factors. Moreover, generalizing emotional tendencies based on attributes such as gender and race can lead to \textit{emotional stereotypes}. For instance, a common stereotype suggests that \textit{men are more prone to anger, whereas women are more likely to express sadness or other emotions} \cite{shields2013gender}. Similarly, emotional stereotypes can also exist specific to nationalities, i.e. \textit{French individuals can be stereotyped as expressing more passion and romantic emotions, while Japanese individuals can be stereotyped as feeling more shame or embarrassment} \cite{mesquita1992cultural}.

 Thus, in this work, we investigate the problem of emotional stereotypes in LLMs when \textit{nationality-based personas} are assigned to the LLM. We seek to answer the following research question: (\textbf{RQ1}) 
 \textit{How do LLMs attribute emotions differently when personas from different nationalities are assigned, and what patterns emerge in these attributions?} 
 At a glance, our experimental framework to undertake this analysis is shown in Figure \ref{fig:example}.

Finally, as LLMs undergo significant alignment training (i.e., RLHF \cite{ouyang2022training, stiennon2020learning}) to align them with human values and ethics, our framework can also uncover whether nationality-specific emotional attribution by LLMs is actually aligned with individuals that belong to those nationalities. We further extend this audit to intersectional cases by examining how nationality and gender jointly influence emotional attribution in LLMs. 
A mismatch in human responses and LLM responses at either the nationality level or the nationality × gender level will indicate the need for improved alignment practices that are better tailored for individuals from different nationalities. Hence, the second research question we investigate in our work is: (\textbf{RQ2}) \textit{How do nationality-specific emotional attributions by LLMs compare to the cultural norms found in these countries?}

In sum, we make the following contributions:
\begin{itemize}[nosep]
    \item To the best of our knowledge, our work is the first to systematically analyze \textit{nationality-specific emotion biases} in LLMs by uncovering differences in LLM emotion attribution across personas from various nationalities.
    \looseness-1\item Our findings reveal statistically significant nationality-based differences in LLM emotion attribution that overgeneralize cultural norms, thereby reinforcing stereotypes. We also observe appreciable misalignment between LLM-generated and human responses, especially for \textit{negative} emotions (e.g., anger). 
    \looseness-1\item We incorporate four core cultural dimensions Power Distance, Uncertainty Avoidance, Long-Term Orientation, and Individualism from Hofstede’s cross-cultural framework to interpret and compare these emotional patterns, providing a richer understanding of how LLMs internalize and reproduce culturally structured affect. 
\end{itemize}







 


\section{Related Work}\vspace{-2mm}

\looseness-1\noindent\textbf{Persona and LLMs.} Many recent studies have worked on persona-based LLMs, where they focused on how assigning different types of persona affects the performance of the LLMs \cite{beck2024sensitivity, mukherjee-etal-2024-cultural, kamruzzaman2024woman}. They found that LLMs are sensitive to assigned personas, with performance varying depending on the specific persona. Some personas improve performance and reduce social bias, while others lead to decreased performance due to inherent biases in LLMs~\cite{gupta2023bias, kamruzzaman2024prompting, de2024helpful}. \vspace{1mm}


\looseness-1\noindent\textbf{Personalization Tradeoffs.} Demographic or role-based personas can boost task utility yet simultaneously introduce bias and safety risks, making persona steering a lively topic in ongoing alignment research. Recent work documents how such prompts modulate toxicity, stereotype uptake, and performance, underscoring the need for techniques that balance personalization with fairness and safety \citep{cheng2023marked,liu2024evaluating,vijjini-etal-2025-exploring}.

\noindent\textbf{Emotion Attribution in LLMs.} Emotion attribution studies found that LLMs exhibit elements of cognitive empathy, such as recognizing emotions and providing emotionally supportive responses across various contexts~\cite{sorin2024large,welivita2024large}. Recent studies have identified gendered and religious emotion attribution in LLMs~\cite{plaza-del-arco-etal-2024-angry, sadhu-etal-2024-empirical,plaza-del-arco-etal-2024-divine}, where they found that emotion attribution changes widely based on gender and religion. \citet{rai-etal-2025-social} offers a distinct perspective by analyzing the portrayal of emotions like shame and pride in cinematic dialogues across cultures. Their study reveals that cultural frameworks influence emotional representation—for instance, Indian films (Bollywood) often depict shame as a collective, socially visible experience, whereas Western films (Hollywood) tend to frame it as an individual, internally experienced emotion—underscoring the cultural construction of emotional significance.


\looseness-1\noindent\textbf{Alignment of LLMs.} Aligning LLMs with human values and expectations is crucial for ensuring their outputs are helpful, truthful, and safe. Recent studies have explored various alignment techniques, including data collection, training methodologies, and evaluation strategies~\cite{wang2023aligning, shen2023large, cao2024towards}. \citet{kirk2024prism} underscores the need to consider multicultural perspectives in alignment, as cultural backgrounds shape interactions with LLMs. Further, \citet{wei2022emergent} argues that alignment must also address emergent behaviors in LLMs, as these models can develop unintended capabilities that are not controllable by conventional safety protocols. 



\section{Experimental Setup}

\looseness-1\textbf{Dataset.} We use the International Survey on Emotion Antecedents and Reactions (ISEAR) \cite{scherer1994evidence} data. ISEAR includes 7,665 events of 7 emotion categories (anger (1,096), fear (1,095), sadness (1,096), joy (1,094), disgust (1,096), guilt (1,093), and shame (1,096)). We utilize information from 3000 respondents in the dataset covering 16 countries. We selected the ISEAR dataset deliberately and thoughtfully for several key reasons. 

First, our study involves analyzing and comparing LLM responses with actual human emotional responses from people across different countries (as part of RQ2). To the best of our knowledge, ISEAR is the only dataset that provides culturally grounded emotional attributions tied to specific events and individuals from diverse national backgrounds. This aspect is crucial to our analysis, as it enables us to evaluate whether LLMs reproduce emotional patterns that are grounded in real-world, cross-cultural human data, a central contribution of our work. 

Second, although the ISEAR dataset’s emotion categories are limited, it is important to note that the emotional events described in the dataset were designed to elicit the full range of those available emotions. That is, the events themselves are emotionally appropriate for the emotion categories provided. For example, the dataset includes scenarios such as conflict, loss, and fear-inducing situations, which naturally align with emotions like anger, sadness, or fear. In other words, we are not forcing models to choose from a skewed emotional set for mismatched scenarios (e.g., joyful events with only negative labels). The event-emotion alignment is a key strength of the ISEAR dataset and helps ensure that our findings are contextually valid despite the limited emotional spectrum.

\vspace{1mm}


\noindent\textbf{Models.}
We use four LLMs in our experiments namely Gemma2-9B, Llama3.2-3B, Mistral-7B, and GPT4o-mini (more details in \Cref{app:model_details}).\vspace{1mm}

\looseness-1\noindent\textbf{Persona Assignment.} We use 110 nationality personas to explore variations in emotional perspectives across different models on a global scale. These personas are based on 110 countries (full list in \Cref{app:all_countires}) recognized by the United Nations (UN).\footnote{\scriptsize{\url{https://www.un.org/dgacm/en/content/regional-groups.}}} The UN organizes their countries into five regions: \textit{Asia-Pacific States, Western European and Other States, Eastern European States, African States,} and \textit{Latin American and Caribbean States}. Since our study uses nationality personas to assess emotional responses, grouping countries by regions provided a structured and interpretable way to compare emotional attributions across a manageable number of personas. The use of regional groupings in our experimental design is both intentional and methodologically grounded. This form of aggregation has also been adopted in related work on cultural variation in cross-cultural NLP research \cite{kamruzzaman-kim-2025-exploring}. 
To ensure equal representation, we select 22 random countries from each region. Since the Eastern European States region has the fewest countries (22), we include all of them. 
This ensures a balanced analysis across all regions and results in a dataset containing 7,665 × 110 = 843,150 examples. We assign LLMs personas using three prompting template variations (see \Cref{app:persona_template}). This ensures our results are robust, as prior work has shown different prompting templates drastically influence LLM outputs \cite{beck2024sensitivity, sclar2023quantifying}. To see more details about response handling, see \Cref{app:abstention}. \vspace{1mm}

\begin{figure*}[t]
\centering
\includegraphics[width=1.0\linewidth]{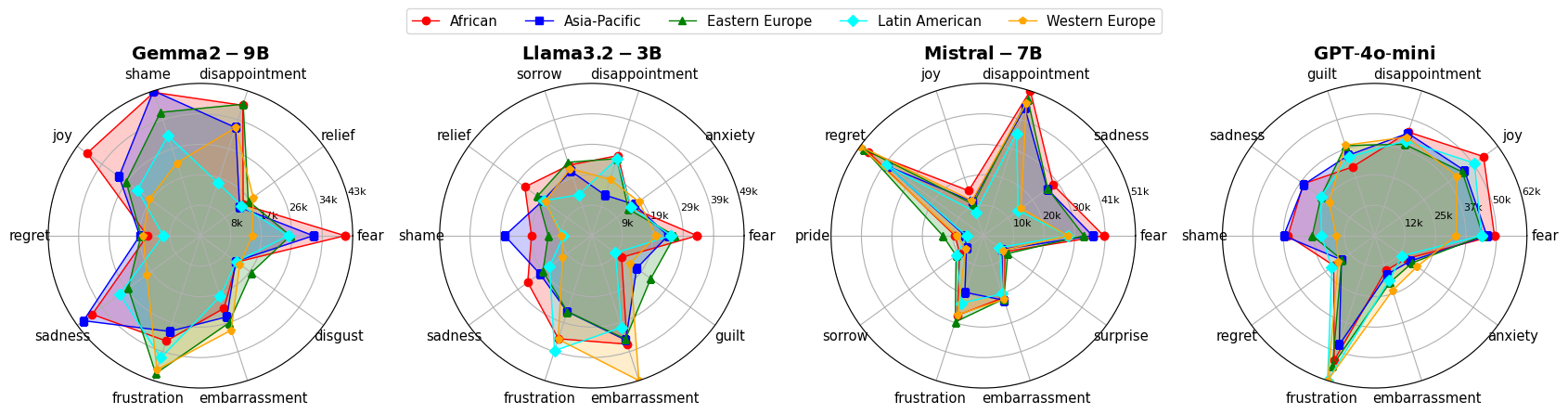}
\caption{The top 10 most frequent emotions by \textbf{region} for each LLM, averaged across all three prompting templates.}
\label{fig:rq1_radar_plot}\vspace{-1.5mm}
\end{figure*}

\section{Results and Discussions}
\label{sec:exp_results}

We first discuss the extensive experiments conducted to answer both RQ1 and RQ2 below, as well as the insights derived from our findings and observations. Before we discuss our results, 
note that our prescriptive stance is that a \textbf{\textit{fair and aligned}} LLM should exhibit behavior that reflects culturally grounded emotional norms while avoiding reductive or stereotypical portrayals of groups based on nationality, culture, or gender. Rather than defaulting to narrow emotional archetypes, a well-aligned model should mirror the diversity of human emotional responses observed within communities, recognizing that even within a given culture emotional expression varies significantly. The aim is not to ignore cultural patterns, but to ensure that models do not collapse rich, heterogeneous emotional norms into oversimplified or homogenized outputs. Essentially, it should align with actual user responses from those communities and cultures. 
The adaptable nature of LLMs introduces an inherent \textit{trade-off} between achieving fairness and enabling personalization.  Ongoing alignment research has recently raised the question of \textit{personalization scale}, i.e. \textit{whether a model should be aligned for a broad ``average user'' or to finely delineated user groups}?~\cite{schwerzmann2025desired,korinek2022aligned,hristova2024problem,gabriel2020artificial}. When LLMs are deployed as general-purpose tools, equitable treatment across diverse groups is paramount; yet when models are customized to mirror specific users, reproducing those users’ idiosyncratic biases may be an intentional (if risky) form of personalization.  Note that our goal in this paper is \textbf{not} to identify this \textit{ideal} alignment scale.  Instead, we seek to provide empirical evidence that current alignment techniques operate at too coarse a level, i.e. \textit{they model the relationship between countries and emotion attributions} at a resolution that collapses meaningful within-country variation. This coarse attribution granularity can over-generalize emotional patterns, leading LLMs to stereotype or misattribute emotions for particular countries.\\

\begin{figure*}[t]
\centering
\includegraphics[width=1.0\linewidth]{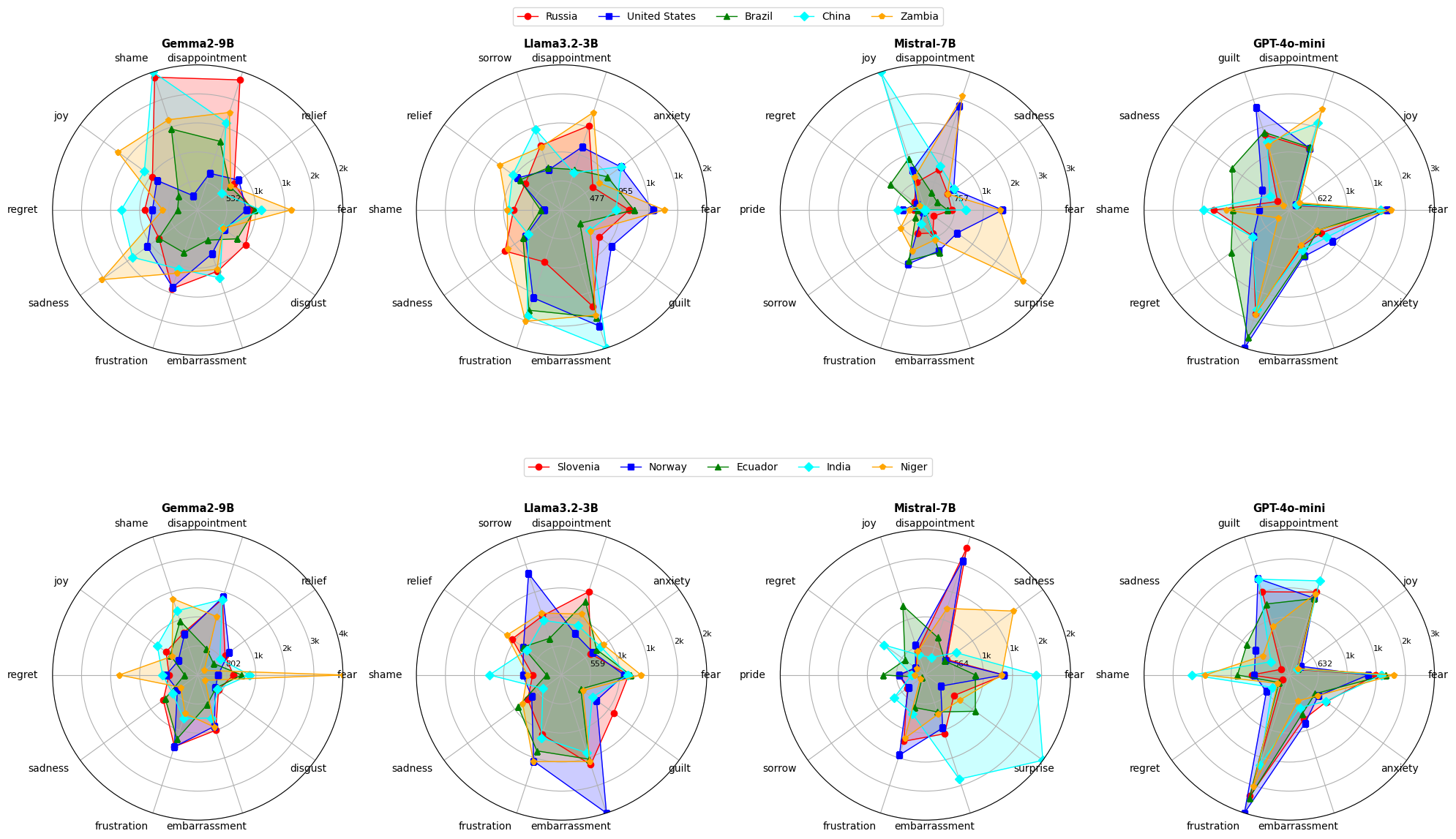}
\caption{The top 10 most frequent emotions for \textbf{10 countries} across each LLM, averaged over all three prompting templates.}
\label{fig:rq1_rcultural_dimension}
\end{figure*}

\noindent \fbox{%
    \parbox{0.455\textwidth}{\textbf{\underline{RQ1}: \textit{How do LLMs attribute emotions differently with personas assigned to different nationalities, and what attributional patterns emerge?}}}}
\label{sec:rq1}
\vspace{3mm}


\noindent In \Cref{fig:rq1_radar_plot}, we present the results for all four models averaged across three prompting templates. We discuss insights from our overall results below. 
For \Cref{sec:subsection1,sec:subsection2,sec:subsection3,sec:subsection4} (emotion attribution across regions and models), we use the task prompt \textit{``What is the main emotion you would
feel while experiencing this event \{event\}? Answer with a single emotion. We don't need explanations for your response"}. 

\subsection{The emotional attribution patterns of LLMs reveal distinct regional disparities} \label{sec:subsection1}\vspace{-1mm}

As shown in \Cref{fig:rq1_radar_plot}, emotion attributions vary significantly across regions. \textit{Shame} is more frequently assigned to Asia-Pacific states, while \textit{fear, joy}, and \textit{disappointment} are predominant for African states. LLMs respond with more \textit{embarrassment} and \textit{regret} for Western European states, while associating \textit{frustration} and \textit{disgust} more with Eastern European states. Most of these regional emotion-attribution differences were statistically significant under
the Chi-squared ($\chi^2$) test \cite{greenwood1996guide} (please refer to \Cref{app:stat} for results). 

\begin{table*}[h]
    \centering
    \caption{Region-wise refusal rates for LLMs, averaged across all the prompting templates. }
    \label{tab:region_refusals}\vspace{-2mm}
    {\small
    \begin{tabular}{lrrrrr}
        \toprule
        \textbf{Model} & \textbf{African} & \textbf{Asia-Pacific} & \textbf{Eastern European} & \textbf{Latin American} & \textbf{Western European} \\
        \midrule
        Lllama3.2-3B  & 18,228 (3.60\%) & 30,279 (5.98\%) & 19,332 (3.82\%) & 16,885 (3.34\%) & 19,295 (3.81\%) \\
        Gemma2-9B  & 2 (0.00\%) & 0 (0.00\%) & 0 (0.00\%) & 1 (0.00\%) & 4 (0.00\%) \\
        Mistral-7B & 30 (0.01\%) & 96 (0.02\%) & 51 (0.01\%) & 12 (0.00\%) & 57 (0.01\%) \\
        GPT-4o-mini  & 0 (0.00\%) & 0 (0.00\%) & 0 (0.00\%) & 0 (0.00\%) & 0 (0.00\%) \\
        \bottomrule
    \end{tabular}
    }
\end{table*}

\subsection{Different LLMs have different emotional attributional patterns}\label{sec:subsection2}\vspace{-1mm} 

From \Cref{fig:rq1_radar_plot}, we can see that the \textit{Gemma} LLM assigns more \textit{shame} and \textit{sadness} but less \textit{frustration} to Asia-Pacific states, while predominantly associating \textit{fear} and \textit{joy} with African states. The \textit{Mistral} LLM attributes \textit{pride} and \textit{frustration} more to Eastern European states and \textit{sorrow} to Latin America. It also mirrors \textit{Gemma} and \textit{Llama} in linking \textit{fear} and \textit{disappointment} more frequently to African states. \textit{Llama} on the other hand, associates higher \textit{frustration} with Latin American states and \textit{embarrassment} with Western European states. The closed-source \textit{GPT-4o-mini} LLM is similar to \textit{Gemma} in its higher attribution of \textit{shame} and \textit{sadness} to Asia-Pacific states and \textit{joy} and \textit{fear} to African states. It also responds with more \textit{embarrassment, guilt}, and \textit{anxiety} for Western states.

\subsection{Analyzing Cultural and Emotional Dimensions through Hierarchy, Uncertainty, and Time Orientation (PDI, UAI, LTO)}\label{sec:subsection3}\vspace{-1mm} 

We perform a cultural-dimension analysis using Hofstede’s country-level framework \cite{hoftede2010cultures}, incorporating Power Distance (PDI), Uncertainty Avoidance (UAI), and Long-Term Orientation (LTO) to interpret emotion patterns across countries. These dimensions capture: (1) acceptance of hierarchy, (2) tolerance for ambiguity, and (3) emphasis on long- versus short-term orientation. We adopt country-level cultural scores from the \textbf{Hofstede Insights/The Culture Factor Country Comparison Tool} \footnote{\scriptsize{\url{https://www.theculturefactor.com/country-comparison-tool}}}, which builds directly on Hofstede’s validated six-dimension model \cite{hoftede2010cultures} with contemporary survey data—one of the most widely used and empirically validated frameworks in cross-cultural psychology. For this analysis, we take two countries per UN region: Russia (PDI = 93, UAI = 95, LTO = 58) \& Slovenia (PDI = 71, UAI = 88, LTO = 50) (Eastern Europe), United States (PDI = 40, UAI = 46, LTO = 50) \& Norway (PDI = 31, UAI = 50, LTO = 55) (Western), Ecuador (PDI = 78, UAI = 67, LTO = 24) \& Brazil (PDI = 69, UAI = 76, LTO = 28) (Latin America), China (PDI = 80, UAI = 30, LTO = 77) \& India (PDI = 77, UAI = 40, LTO = 51) (Asia-Pacific), and Zambia (PDI = 60, UAI = 50, LTO = 30) \& Niger (PDI = 80, UAI = 55, LTO = 8) (Africa). In \Cref{fig:rq1_rcultural_dimension}, we present the country-wise results for these 10 countries. Across the four models, consistent yet model-specific cultural patterns emerge. Gemma2-9B tends to align most closely with Hofstede’s framework, associating hierarchical cultures with self-conscious emotions such as shame and fear and egalitarian ones with more positive affect like joy and relief. Llama3.2-3B shows a similar hierarchy effect but diverges in uncertainty avoidance, depicting rule-bound societies as calmer rather than more anxious. GPT-4o-mini mirrors these general tendencies but with weaker correlations, mapping high-hierarchy contexts to shame and fear and low-hierarchy ones to frustration and embarrassment, while displaying only mild links to uncertainty or time orientation. Mistral 7B follows Hofstede-like cues through different emotional proxies—embarrassment and surprise replacing shame—suggesting that cultural bias manifests differently across model architectures. Together, these findings indicate that while all models reproduce recognizable cultural heuristics, their emotional mappings vary in direction and intensity, reflecting model-specific pathways through which cultural stereotypes are encoded. For more detailed discussion see \Cref{app:details}. 

\subsection{LLM refusal rates vary across regions}\label{sec:subsection4}

Although we instructed the LLMs to respond with a single emotion, the models abstain from answering questions at times. In \Cref{tab:region_refusals}, we present the refusal rates of each model, expressed as the percentage of refusals along with the exact number of instances. We observe that the Llama3.2 model exhibits the highest refusal rate across most cases, whereas other models have fewer refusals, with GPT4o-mini showing no refusals at all. 
Additionally, for the same events but different persona assignments, we observe that the refusal rate is higher for the Asia-Pacific region than for any other region, with certain countries, such as North Korea, Saudi Arabia, Iraq, Afghanistan, and Ukraine, contributing disproportionately to these refusals. Models generally refuse to answer family- or friendship-related events in these countries as well as any content that could be potentially suggestive. We provide a detailed breakdown of Llama3.2 country-wise refusal rates for Llama3.2 in \Cref{app:abstention}.


\vspace{1mm}
\noindent\underline{\textbf{Overall Discussion on RQ1.}} From the above results, we see that LLMs exhibit distinct regional disparities in emotional attributions when assigned nationality-specific personas. These disparities are particularly significant as they risk reinforcing harmful societal stereotypes, contributing to the undesirable LLM impacts
known as \textbf{\textit{representational harm}} \cite{blodgett-etal-2020-language}. Representational harm occurs when groups are depicted in reductive or stereotypical ways, potentially limiting or misrepresenting the diversity of individuals within those groups. For instance, consistent attribution of \textit{shame} to Asia-Pacific states and \textit{fear} or \textit{joy} predominantly to African states may reinforce simplified and culturally reductive narratives. Such portrayals, while possibly reflecting common stereotypes, overlook nuanced emotional diversity within these cultures, leading to misrepresentation and oversimplification of complex emotional norms. 

Another notable observation from our analysis pertains to the variability in refusal rates across regions, with notably higher rates for the Llama3.2 model. LLMs showed significantly elevated refusal rates for prompts associated with countries like North Korea and Saudi Arabia. This might stem from the sensitive nature of the prompts and potential over-cautiousness programmed into alignment mechanisms within the model. Particularly, for North Korea, the model frequently abstained even from benign life-experience scenarios or universally positive emotional events, which \textit{suggests heightened sensitivity or possibly overly restrictive content filters for certain nationalities}.
Next, we discuss RQ2 and related experiments/results.




\begin{table*}[h!]
\centering
{
\fontsize{9pt}{9pt}\selectfont
\caption{Llama3.2-3B generated emotions and explanations, given an event. The LLM makes many stereotypical assumptions-- for instance, about Indians, stating that \textit{heated discussions would be common due to cultural and family expectations.} This might not depict how individuals view their cultural values and can reinforce stereotypes.}\label{tab:gendered_emotions}\vspace{-2mm}
\begin{tabular}{p{1.5cm}p{13cm}}
\hline
\textbf{Event} & ``Heated discussion with spouse concerning new house.'' \\
\hline
\textit{China} & \textbf{Shame.} ``As a Chinese person, I value \textbf{\textit{harmony}} and respect for elders. In a heated discussion with my spouse about our new house, I would feel shame because it implies that we cannot work together to resolve our differences, which \textbf{\textit{goes against the Confucian values of familial unity and cooperation.}}'' \\
\hline
\textit{India} &
\textbf{Frustration.} ``In India, I would \textbf{\textit{expect heated discussions to be common due to the many cultural and family expectations surrounding marriage and home life}}. My spouse might disagree on various aspects of our new house, like decor, location, or budget. I would feel frustrated if they raise their voice while discussing it because in \textbf{\textit{Indian culture, maintaining peace is paramount, especially when it comes to matters concerning a partner's feelings.}}'' \\
\hline
\textit{United States} & \textbf{Anxiety.} ``I'd choose anxiety because I'd be worried about how the conversation might \textbf{\textit{escalate or affect our relationship, and uncertainty about the outcome would make me feel uneasy and on edge.}}'' \\
\hline
\end{tabular}
}
\end{table*}

\begin{table}[ht]
\centering
{
\fontsize{6pt}{6pt}\selectfont
\caption{Accuracy scores for \textit{countries} by \textit{emotions}, averaged across all LLMs. The \textcolor{blue} {\textbf{blue}} color indicates the highest accuracy in that particular emotion and \textcolor{red} {\textbf{red}} indicates the lowest accuracy of that particular emotion.}\label{tab:accuracy_countries_emotions}\vspace{-2mm}
\setlength{\tabcolsep}{2.5pt}
\resizebox{0.48\textwidth}{!}{
\begin{tabular}{llrccccc}
\hline
\textbf{Country} & \textbf{Joy} & \textbf{Fear} & \textbf{Anger} & \textbf{Sadness} & \textbf{Disgust} & \textbf{Shame} & \textbf{Guilt} \\  
\hline
Bulgaria         & 0.93         & 0.76         & 0.59         & 0.91         & 0.59         & 0.43         & 0.74         \\ 
Sweden           & 0.94         & 0.70         & \textcolor{red} {\textbf{0.33}}         & 0.76         & 0.49         & \textcolor{red} {\textbf{0.21}}         & 0.62         \\ 
Norway           & 0.89         & 0.65         & 0.45         & 0.93         & 0.32         & 0.25         & 0.67         \\ 
Finland          & 0.94         & 0.67         & 0.43         & 0.92         & 0.60         & 0.32         & 0.70         \\ 
Austria          & 0.91         & \textcolor{blue} {\textbf{0.82}}         & 0.45         & \textcolor{blue} {\textbf{0.95}}        & 0.62         & 0.49         & 0.73         \\ 
Australia        & \textcolor{red} {\textbf{0.85}}         & 0.67         & 0.46         & 0.92         & 0.46         & 0.20         & \textcolor{blue} {\textbf{0.87}}        \\ 
N. Zealand      & 0.95         & 0.68         & 0.44         & 0.80         & 0.55         & 0.30         & 0.80         \\ 
Netherlands      & 0.88         & 0.70         & 0.43         & 0.91         & 0.57         & 0.38         & 0.73         \\ 
Spain            & 0.91         & 0.69         & 0.41         & 0.87         & \textcolor{blue} {\textbf{0.70}}        & 0.42         & 0.60         \\ 
USA              & 0.95         & 0.73         & 0.65         & 0.87         & 0.41         & 0.20         & 0.79         \\ 
Brazil           & 0.94         & 0.72         & 0.56         & 0.88         & 0.61         & 0.55         & 0.64         \\ 
Honduras         & 0.91         & 0.73         & 0.60         & 0.82         & 0.57         & 0.54         & 0.83         \\ 
India            & 0.97         & 0.69         & 0.55         & 0.93         & 0.47         & 0.50         & 0.61         \\ 
China   & 0.93         & \textcolor{red} {\textbf{0.59}}         & 0.36         & \textcolor{red} {\textbf{0.75}}         & 0.41         & 0.34         & \textcolor{red} {\textbf{0.47}}         \\ 
Zambia           & \textcolor{blue}{\textbf{0.98}}        & 0.67         & 0.56         & 0.92         & \textcolor{red} {\textbf{0.31}}         & 0.39         & 0.62         \\ 
Malawi           & 0.97         & 0.73         & \textcolor{blue} {\textbf{0.66}}        & 0.89         & 0.53         & \textcolor{blue} {\textbf{0.65}}        & 0.78         \\ \hline
\end{tabular}}
}\vspace{-6mm}

\end{table}

\vspace{4mm}
\noindent \fbox{%
    \parbox{0.455\textwidth}{\textbf{\underline{RQ2}: \textit{How do nationality-specific emotional attributions by LLMs compare to the cultural norms found in these countries?}}}}\\

In subsequent subsections, we detail our efforts and findings in assessing whether LLM responses align adequately with those from humans from specific nationalities and countries. 




\subsection{Analyzing Cultural and Emotional Dimensions through Individualism} 
To answer RQ2, we first undertake a qualitative analysis. We randomly select 2000 examples from the dataset (4 countries from each region) and prompt the LLMs to generate an emotion as well as a rationale for their choice. We use the following task prompt: \textit{``What is the main emotion you would feel while experiencing this event: \{event\}? Answer with a single emotion and provide a short explanation''}. We then manually reviewed a subset of responses and identified recurring patterns in how emotions were expressed across countries. Some illustrative examples are presented in \Cref{tab:gendered_emotions} and \Cref{app:model_explanations}. To systematically analyze these trends at scale, we further applied topic modeling using BERTTopic \cite{grootendorst2022bertopic} on the obtained LLM rationales, allowing us to uncover dominant themes in the responses, complementing our manual review. The generated topics are provided in \Cref{app:topic_modeling}.

\looseness-1As seen in our qualitative results, many responses echo a well‐established dichotomy from Hofstede’s \textit{cultural dimensions}~\cite{hofstede1980culture,markus1991cultural}, on \textit{individualism} vs. \textit{collectivism}:

\begin{itemize}[nosep]
    \item \textit{Collectivistic Societies} (e.g., China, India, Cuba, Malawi): Emphasis on social harmony, family expectations, concern about bringing dishonor to the group.
    \item \textit{Individualistic Societies} (e.g., US, UK): Emphasis on personal freedom, individual emotional well‐being, or personal rights.
\end{itemize}

\looseness-1This is also observable in \Cref{tab:gendered_emotions}, as Llama-3.2 associates emotions with these cultural values (e.g., \textit{China} $\rightarrow$ \textit{Harmony}, \textit{India} $\rightarrow$ \textit{Collective Family Obligations}, \textit{US} $\rightarrow$ \textit{Individual Concerns}).
%
The fact that LLMs default to these themes may indeed reflect an alignment with the broad contours of cultural psychology, but this alignment often comes at the cost of overlooking the rich diversity of individual experiences and expressions~\cite{eid2001norms}. Some responses also appear to recite textbook values---e.g., references to \textit{Confucian harmony}~\cite{li2006confucian}; \textit{family is paramount}~\cite{hofstede2001culture,markus2014culture}, etc. While this can be valuable if it reflects accurate discourses, it can also reinforce stereotypes and fail to capture how real individuals might deviate from these norms in practice. One illustrative case of stereotypical assumption appears in the Llama3.2-3B response for the Indian persona (as in \Cref{tab:gendered_emotions}). When presented with the event \textit{``Heated discussion with spouse concerning new house''}, the model responds with frustration, and justifies it by stating that `heated discussions are common in India due to cultural and family expectations'. While the surface rationale appears aligned with collectivistic cultural tropes often associated with Indian society such as the importance of extended family and marital expectations this type of response exhibits problematic tendencies. The model appears to default to a scripted narrative where Indian families are assumed to have emotionally charged, expectation-laden interactions over household decisions implicitly framing such conflict as `normal' or even inevitable. By asserting that such discussions are expected in Indian households, the model risks reinforcing essentialist and deterministic assumptions about Indian emotional norms. This mirrors a broader trend we observed in LLM outputs, where emotional attributions are less grounded in real individual perspectives and more aligned with overgeneralized cultural archetypes. So, while the LLMs' outputs show some surface‐level alignment with cultures, they rely on overgeneralizations and stereotypes. Next, we discuss whether these responses align with those from actual individuals belonging to specific nationalities. 

\subsection{Assessing LLM alignment with actual nationality-specific human responses}
We now prompt the models to respond within the seven emotions as in the ISEAR dataset for a quantitative comparison with humans. We use the same prompting templates as listed in \Cref{tab:prompting_templates}, and modify the task prompt slightly: \textit{``What is the main emotion you would
feel while experiencing this event \{event\}? Answer with one of the following emotions: anger, fear, sadness, joy, disgust, guilt, or shame. We don't need explanations for your response"}. We discuss our insights below.

\looseness-1In Table \ref{tab:accuracy_countries_emotions}, we present the accuracy results averaged across all LLMs.\footnote{Model-wise results \& F1-scores are in Appendix \ref{app:model-wise}.} 
Most countries show high accuracies for \textit{joy} (e.g., Zambia at 0.98 accuracy) and for \textit{sadness} (e.g., Austria at 0.95). This suggests that, across countries, LLMs handle these two emotions fairly reliably. Emotions such as \textit{anger, disgust, shame}, and \textit{guilt} vary widely. For instance, \textit{anger} ranges from as low as 0.33 (Sweden) to as high as 0.66 (Malawi); \textit{disgust} ranges from 0.31 (Zambia) to 0.70 (Spain); and \textit{shame} varies from 0.20 (Australia) to 0.65 (Malawi). In sum, our findings indicate accurate alignment with \textit{joy} and \textit{sadness}, and inaccurate alignment with other negative emotions.

\looseness-1These swings point to challenges LLMs face in capturing subtler or more culturally nuanced negative emotions. Our results demonstrate the need for improved LLM alignment strategies either from the data or methodology perspective. LLMs are not aligned well with human users from varying nationalities when it comes to nuanced emotion attribution, potentially showcasing gaps in their understanding of nuanced societal and cultural norms across these different countries. Along the same lines, we will now study whether providing additional gender-related information improves alignment or leads to further overgeneralization.


\subsection{Intersectionality analysis using additional gender-related information}
\label{sec:interseconality}

\begin{figure*}[t]
\centering
\includegraphics[width=0.89\linewidth]{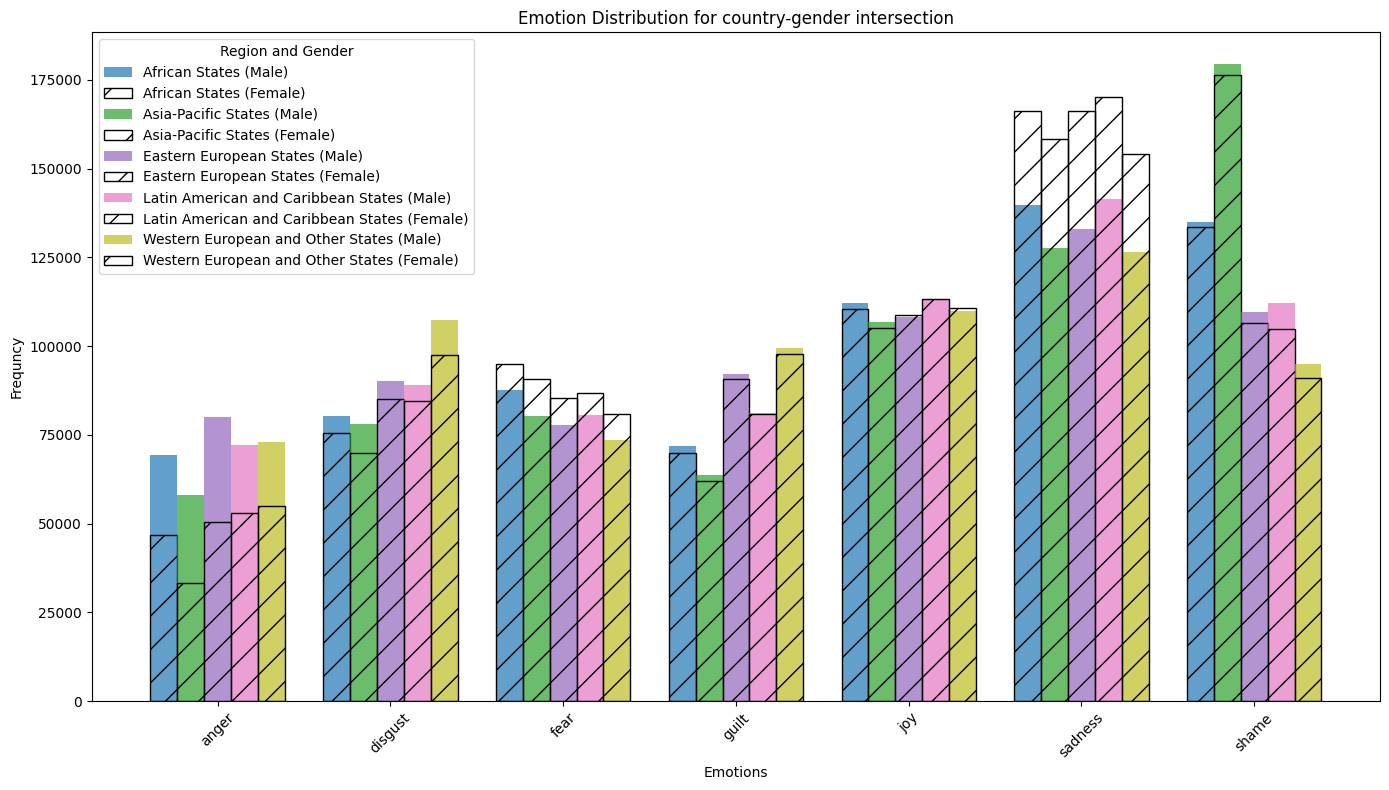}\vspace{-3mm}
\caption{Results for intersectional experiments for five regions averaged across all four models and prompting templates. The results reveal notable region and gender disparities in emotion attribution. Specifically, LLMs tend to associate \textit{anger} and \textit{shame} more frequently with \textit{male-country}, and \textit{sadness} and \textit{fear} to \textit{female-country} intersections. }\vspace{-3mm}
\label{fig:intersectional_bias}
\end{figure*}

Since ISEAR also possesses gender-related information from real human participants, we conduct an intersectionality analysis by providing LLMs with gender-related attributes, along with the original nationality-specific persona. To do this, we include both \textit{gender} and \textit{country} sociodemographic information in the task prompt, similarly localized to the seven emotions for the human participant responses. Our prompts thus change only slightly. For example, consider the P1 persona from \Cref{tab:prompting_templates}, which now becomes: \textit{`You are a \{gender\} from \{persona\}. Your responses should closely mirror the knowledge and abilities of this
persona.'} Here, the\textit{ \{gender\}} attribute can take on either `a male' or `a female' persona as in the original ISEAR dataset survey responses from humans.

\definecolor{malerow}{RGB}{240,240,240} 

\begin{table}[ht]
\centering
\fontsize{6pt}{6pt}\selectfont
\caption{Accuracy scores for \textit{gender--country} by \textit{emotions}, averaged across all LLMs. The \textcolor{blue}{\textbf{blue}} color indicates the highest accuracy in that particular emotion and \textcolor{red}{\textbf{red}} indicates the lowest accuracy of that particular emotion. `M' $\rightarrow$ `Male' and `F' $\rightarrow$ `Female'.}
\label{tab:accuracy_gender_country_emotions}\vspace{-2mm}
\setlength{\tabcolsep}{2.5pt}
\rowcolors{2}{malerow}{white}   
\resizebox{0.48\textwidth}{!}{%
\begin{tabular}{llrccccc}
\hline
\textbf{Country--Gender} & \textbf{Joy} & \textbf{Fear} & \textbf{Anger} & \textbf{Sadness} & \textbf{Disgust} & \textbf{Shame} & \textbf{Guilt} \\  
\hline
Bulgaria--M   & 0.90 & 0.75 & 0.64 & 0.92 & 0.64 & 0.44 & 0.69 \\ 
Bulgaria--F   & 0.90 & 0.81 & 0.58 & 0.96 & 0.58 & 0.44 & 0.69 \\ 
Sweden--M     & 0.91 & 0.69 & 0.38 & 0.77 & 0.54 & 0.22 & 0.57 \\ 
Sweden--F     & 0.91 & 0.75 & 0.41 & 0.81 & 0.48 & 0.22 & 0.57 \\ 
Norway--M     & 0.86 & 0.64 & 0.50 & 0.94 & 0.37 & 0.26 & 0.62 \\ 
Norway--F     & 0.86 & 0.70 & 0.44 & 0.98 & 0.31 & 0.26 & \textcolor{blue}{\textbf{0.82}} \\ 
Finland--M    & 0.91 & 0.66 & 0.48 & 0.93 & 0.65 & 0.33 & 0.65 \\ 
Finland--F    & 0.91 & 0.72 & 0.42 & 0.97 & 0.59 & 0.33 & 0.65 \\ 
Austria--M    & 0.88 & 0.81 & 0.50 & 0.96 & \textcolor{blue}{\textbf{0.76}} & 0.50 & 0.68 \\ 
Austria--F    & 0.88 & \textcolor{blue}{\textbf{0.87}} & 0.44 & \textcolor{blue}{\textbf{0.99}} & 0.61 & 0.50 & 0.68 \\ 
Australia--M  & \textcolor{red}{\textbf{0.80}} & 0.66 & 0.51 & 0.93 & 0.51 & 0.26 & 0.80 \\ 
Australia--F  & 0.82 & 0.72 & 0.45 & 0.97 & 0.45 & 0.28 & 0.62 \\ 
N. Zealand--M & 0.92 & 0.67 & 0.49 & 0.81 & 0.60 & \textcolor{red}{\textbf{0.21}} & 0.75 \\ 
N. Zealand--F & 0.92 & 0.73 & 0.43 & 0.85 & 0.54 & 0.31 & 0.75 \\ 
Netherlands--M & 0.85 & 0.69 & 0.48 & 0.92 & 0.62 & 0.39 & 0.68 \\ 
Netherlands--F & 0.85 & 0.75 & 0.42 & 0.96 & 0.56 & 0.39 & 0.68 \\ 
Spain--M      & 0.88 & 0.68 & 0.46 & 0.88 & 0.75 & 0.43 & 0.55 \\ 
Spain--F      & 0.88 & 0.74 & \textcolor{red}{\textbf{0.32}} & 0.92 & 0.69 & 0.43 & 0.55 \\ 
USA--M        & 0.92 & 0.72 & 0.70 & 0.88 & 0.46 & 0.24 & 0.74 \\ 
USA--F        & 0.92 & 0.78 & 0.64 & 0.92 & 0.40 & 0.23 & 0.74 \\ 
Brazil--M     & 0.91 & 0.71 & 0.61 & 0.89 & 0.66 & 0.56 & 0.59 \\ 
Brazil--F     & 0.91 & 0.77 & 0.55 & 0.93 & 0.60 & 0.56 & 0.59 \\ 
Honduras--M   & 0.88 & 0.72 & 0.65 & 0.83 & 0.62 & 0.55 & 0.78 \\ 
Honduras--F   & 0.88 & 0.78 & 0.59 & 0.87 & 0.56 & 0.55 & 0.78 \\ 
India--M      & 0.94 & 0.68 & 0.60 & 0.94 & 0.52 & 0.51 & 0.56 \\ 
India--F      & 0.94 & 0.74 & 0.54 & 0.98 & 0.46 & 0.51 & 0.56 \\ 
China--M      & 0.90 & \textcolor{red}{\textbf{0.58}} & 0.41 & \textcolor{red}{\textbf{0.76}} & 0.46 & 0.35 & 0.42 \\ 
China--F      & 0.90 & 0.64 & 0.35 & 0.80 & 0.40 & 0.35 & \textcolor{red}{\textbf{0.39}} \\ 
Zambia--M     & \textcolor{blue}{\textbf{0.95}} & 0.66 & 0.61 & 0.93 & 0.36 & 0.40 & 0.57 \\ 
Zambia--F     & 0.93 & 0.72 & 0.55 & 0.97 & \textcolor{red}{\textbf{0.30}} & 0.40 & 0.57 \\ 
Malawi--M     & 0.94 & 0.72 & \textcolor{blue}{\textbf{0.78}} & 0.90 & 0.58 & 0.64 & 0.73 \\ 
Malawi--F     & 0.94 & 0.78 & 0.65 & 0.94 & 0.52 & \textcolor{blue}{\textbf{0.66}} & 0.73 \\ 
\hline
\end{tabular}}
\vspace{-4mm}
\end{table}


Adding the gender dimension yields the largest gains in \textit{anger} and \textit{disgust} for \textit{country-male} instances and in \textit{sadness} and \textit{fear} for \textit{country-female} instances. We presented the results for \textit{country-gender} in \Cref{tab:accuracy_gender_country_emotions} and compared the results with the \textit{country} only version as in \Cref{tab:accuracy_countries_emotions}. For example, Malawi–Male reaches an accuracy of 0.78 on \textit{anger} (↑ +0.12 over country-only) and Austria–Male attains 0.76 on \textit{disgust} (↑ +0.14). On the female side, Austria–Female improves to 0.87 on \textit{fear} and 0.99 on \textit{sadness} (both ↑ +0.05–0.06). These shifts align with well-documented stereotypes that associate men with externalising emotions such as \textit{anger/disgust} and women with internalising emotions such as \textit{sadness/fear} \cite{plaza-del-arco-etal-2024-angry}. Conversely, accuracies on \textit{joy} and \textit{guilt} fall almost uniformly after intersectional tagging. The clearest drops appear for Australia–Male \textit{joy} (0.80 vs. 0.85 baseline) and for China-female on \textit{guilt} (0.39 vs. 0.47). Other emotion categories show only marginal movement ( ±0.03), \textit{indicating that the added gender information does not substantially change the models’ predictions outside the stereotype-consistent cases}. Comparing with the country-only version, improvements are concentrated in the four stereotype-linked emotions noted above, while the reductions in \textit{joy} and \textit{guilt} offset some of the benefit. These results show that intersectionality and additional gender-related information sharpen predictions when the additional attribute reinforces prevalent emotion stereotypes, yet offers limited value elsewhere.


We also visualize the results of our intersectional experiments in \Cref{fig:intersectional_bias}, which reveal clear indications of region and gender bias in emotion attribution. Specifically, we observe that when using the \textit{male-country} intersection, the models are more likely to respond with emotions such as \textit{anger} and \textit{shame} compared to the \textit{female-country} intersection. Conversely, the models assign emotions like \textit{sadness} and \textit{fear} more frequently to the female-country intersection than to the male one. These results are statistically significant (see \Cref{tab:statistical_test_without restrictions_gender}).
\looseness-1Notably, the models' tendency to associate \textit{anger} and \textit{shame} with males and \textit{sadness} and \textit{fear} with females aligns with prior findings on gender stereotypes in emotion attribution, where \textit{anger} is often associated with males and \textit{sadness} with females \cite{plaza-del-arco-etal-2024-angry}. Furthermore, we observe a similar regional bias in emotion attribution as discussed in \Cref{sec:subsection1,sec:subsection2}, where the models attribute \textit{shame} more frequently to Asia-Pacific regions. Thus, the results presented in \Cref{fig:intersectional_bias} reflect a combination of both gender and regional biases. Interestingly, we find that the emotion \textit{joy} is assigned similarly across gender and region intersections, which serves as an example of a desired and unbiased response from the models. 

\vspace{1.5mm}
\noindent\underline{\textbf{Overall Discussion on RQ2.}}
Our RQ2 analyses studied whether the emotions that LLMs attribute to nationality-specific personas actually align with culturally grounded human data. The evidence points to only partial alignment.
Qualitative inspection shows that models reproduce cultural dichotomies, as codified in Hofstede’s collectivism/individualism framework. Moreover, the rationales indicate that this form of textbook alignment risks reinforcing stereotypical narratives rather than reflecting the diversity of lived experiences within each country. Furthermore, when we compare model outputs against ISEAR ground-truth labels, agreement is high for \textit{joy} and \textit{sadness} but not for other negative emotions like \textit{anger}, \textit{disgust}, and \textit{guilt}, suggesting that current model alignment is unable to capture subtler, context-dependent signals in text. Augmenting the persona with intersectional gender information reinforces stereotypes as well: \textit{anger} and \textit{disgust} predictions for \textit{male-country} pairs increase, while \textit{sadness} and \textit{fear} predictions for \textit{female-country} pairs increase.

\section{Conclusion}

\looseness-1 We investigated LLMs' emotional attribution patterns when assigned nationality-specific personas. Our findings reveal that LLMs exhibit significant nationality-based biases in emotion attribution, often reinforcing cultural stereotypes. This misalignment is particularly evident for negative emotions like \textit{anger}, \textit{disgust}, and \textit{shame}. Our work underscores the need for more sophisticated alignment strategies that consider the diversity of human emotional responses across diverse cultural contexts. Model-level comparisons show that both open-source and closed-source LLMs over-generalize in systematically different ways. Region-wise and intersectional experiments further reveal that model decisions such as refusing to answer prompts from sensitive countries or defaulting to collectivist/individualist tropes, can cause representational harm, and amplify undesirable biases. 


\section{Limitations}


\noindent\textbf{More Datasets and Countries.} Our analysis covers 110 countries, which is a substantial sample but falls short of the 193 countries recognized by the United Nations. Additionally, although the ISEAR dataset we used includes a robust volume of samples, incorporating additional data sources could enhance the generalizability of our findings to a wider range of nations and emotional expressions.\vspace{2mm}

\noindent\textbf{Analyzing Languages Beyond English.} Our study is limited to English-language datasets and prompts, as we instructed the LLMs to respond exclusively in English. This approach excludes potential variations in emotional attribution that might emerge when models operate in other languages. Future work could explore multilingual datasets and prompts to better understand how language influences emotional expression in LLMs.\vspace{2mm}


\looseness-1\noindent\textbf{Sub-regional Experiments.} In our experiments, we include 110 countries but do not explore sub-national or sub-regional variations. Emotions may vary within a single country or nation due to cultural, linguistic, or socio-economic differences. Incorporating sub-regional personas in future work could provide a more nuanced understanding of these variations. However, this is not a trivial task, given there is a scarcity of datasets that link emotions to intra-country and regional variations.\vspace{2mm}

\looseness-1\noindent\textbf{Toward Broader Intersectional Analysis.}
While this study incorporates one intersectional dimension—gender and country, many other identity intersections remain unexplored. Our focus on gender–country personas represents progress toward more nuanced identity modeling, yet it omits critical combinations such as country–religion, socio-economic status, or multi-trait intersections like gender–religion–nationality. Future work should examine these richer identity configurations to better assess how LLMs handle the complexity of human identity and demographic diversity.\vspace{2mm} 

\noindent\textbf{LLMs.} While our study includes four LLMs, encompassing both open-source and closed-source models, the rapidly growing landscape of LLMs means that our selection may not capture the full spectrum of biases present across all architectures.


\section{Ethics Statement}

This study examines the presence of nationality-based emotional stereotypes in LLMs and their potential misalignment with human emotional expressions. Our research adheres to ethical guidelines by ensuring that no personally identifiable information is used, and all data sources originate from publicly available datasets, such as the ISEAR dataset. We acknowledge that LLMs may reinforce biases present in their training data, and our findings highlight the necessity of improving bias mitigation strategies to enhance fairness and inclusivity in AI-generated responses. Our study does not aim to perpetuate or reinforce stereotypes but rather to expose and analyze their presence in LLMs. We recognize the potential risks of cultural generalization and have taken steps to present findings responsibly, avoiding deterministic claims about national emotional tendencies. Additionally, all experimental procedures were conducted with transparency, and the results are shared to encourage further research on mitigating biases in LLMs.

\section*{Acknowledgements}
This project was fully supported by the University of South Florida.

\bibliography{custom}

@inproceedings{gupta2023bias,
  title={Bias Runs Deep: Implicit Reasoning Biases in Persona-Assigned LLMs},
  author={Gupta, Shashank and Shrivastava, Vaishnavi and Deshpande, Ameet and Kalyan, Ashwin and Clark, Peter and Sabharwal, Ashish and Khot, Tushar},
  booktitle={The Twelfth International Conference on Learning Representations},
  year={2023}
}

@inproceedings{cheng2023marked,
  title={Marked Personas: Using Natural Language Prompts to Measure Stereotypes in Language Models},
  author={Cheng, Myra and Durmus, Esin and Jurafsky, Dan},
  booktitle={Proceedings of the 61st Annual Meeting of the Association for Computational Linguistics (Volume 1: Long Papers)},
  pages={1504--1532},
  year={2023}
}

@article{kamruzzaman2024prompting,
  title={Prompting Techniques for Reducing Social Bias in LLMs through System 1 and System 2 Cognitive Processes},
  author={Kamruzzaman, Mahammed and Kim, Gene Louis},
  journal={arXiv preprint arXiv:2404.17218},
  year={2024}
}

@article{liu2024evaluating,
  title={Evaluating Large Language Model Biases in Persona-Steered Generation},
  author={Liu, Andy and Diab, Mona and Fried, Daniel},
  journal={arXiv preprint arXiv:2405.20253},
  year={2024}
}

@article{scherer1994evidence,
  title={Evidence for universality and cultural variation of differential emotion response patterning.},
  author={Scherer, Klaus R and Wallbott, Harald G},
  journal={Journal of personality and social psychology},
  volume={66},
  number={2},
  pages={310},
  year={1994},
  publisher={American Psychological Association}
}

@article{sclar2023quantifying,
  title={Quantifying Language Models' Sensitivity to Spurious Features in Prompt Design or: How I learned to start worrying about prompt formatting},
  author={Sclar, Melanie and Choi, Yejin and Tsvetkov, Yulia and Suhr, Alane},
  journal={arXiv preprint arXiv:2310.11324},
  year={2023}
}

@inproceedings{beck2024sensitivity,
  title={Sensitivity, performance, robustness: Deconstructing the effect of sociodemographic prompting},
  author={Beck, Tilman and Schuff, Hendrik and Lauscher, Anne and Gurevych, Iryna},
  booktitle={Proceedings of the 18th Conference of the European Chapter of the Association for Computational Linguistics (Volume 1: Long Papers)},
  pages={2589--2615},
  year={2024}
}

@article{kamruzzaman2024woman,
  title={" A Woman is More Culturally Knowledgeable than A Man?": The Effect of Personas on Cultural Norm Interpretation in LLMs},
  author={Kamruzzaman, Mahammed and Nguyen, Hieu and Hassan, Nazmul and Kim, Gene Louis},
  journal={arXiv preprint arXiv:2409.11636},
  year={2024}
}

@article{shields2013gender,
  title={Gender and emotion: What we think we know, what we need to know, and why it matters},
  author={Shields, Stephanie A},
  journal={Psychology of Women Quarterly},
  volume={37},
  number={4},
  pages={423--435},
  year={2013},
  publisher={Sage Publications Sage CA: Los Angeles, CA}
}

@article{de2024helpful,
  title={Helpful assistant or fruitful facilitator? Investigating how personas affect language model behavior},
  author={de Araujo, Pedro Henrique Luz and Roth, Benjamin},
  journal={arXiv preprint arXiv:2407.02099},
  year={2024}
}

@article{dubey2024llama,
  title={The llama 3 herd of models},
  author={Dubey, Abhimanyu and Jauhri, Abhinav and Pandey, Abhinav and Kadian, Abhishek and Al-Dahle, Ahmad and Letman, Aiesha and Mathur, Akhil and Schelten, Alan and Yang, Amy and Fan, Angela and others},
  journal={arXiv preprint arXiv:2407.21783},
  year={2024}
}

@article{team2024gemma,
  title={Gemma 2: Improving open language models at a practical size},
  author={Team, Gemma and Riviere, Morgane and Pathak, Shreya and Sessa, Pier Giuseppe and Hardin, Cassidy and Bhupatiraju, Surya and Hussenot, L{\'e}onard and Mesnard, Thomas and Shahriari, Bobak and Ram{\'e}, Alexandre and others},
  journal={arXiv preprint arXiv:2408.00118},
  year={2024}
}

@article{jiang2023mistral,
  title={Mistral 7B},
  author={Jiang, Albert Q and Sablayrolles, Alexandre and Mensch, Arthur and Bamford, Chris and Chaplot, Devendra Singh and Casas, Diego de las and Bressand, Florian and Lengyel, Gianna and Lample, Guillaume and Saulnier, Lucile and others},
  journal={arXiv preprint arXiv:2310.06825},
  year={2023}
}

@book{greenwood1996guide,
  title={A guide to chi-squared testing},
  author={Greenwood, Priscilla E and Nikulin, Michael S},
  volume={280},
  year={1996},
  publisher={John Wiley \& Sons}
}

@inproceedings{mukherjee-etal-2024-cultural,
    title = "Cultural Conditioning or Placebo? On the Effectiveness of Socio-Demographic Prompting",
    author = "Mukherjee, Sagnik  and
      Adilazuarda, Muhammad Farid  and
      Sitaram, Sunayana  and
      Bali, Kalika  and
      Aji, Alham Fikri  and
      Choudhury, Monojit",
    editor = "Al-Onaizan, Yaser  and
      Bansal, Mohit  and
      Chen, Yun-Nung",
    booktitle = "Proceedings of the 2024 Conference on Empirical Methods in Natural Language Processing",
    month = nov,
    year = "2024",
    address = "Miami, Florida, USA",
    publisher = "Association for Computational Linguistics",
    url = "https://aclanthology.org/2024.emnlp-main.884/",
    doi = "10.18653/v1/2024.emnlp-main.884",
    pages = "15811--15837"
}

@inproceedings{plaza-del-arco-etal-2024-angry,
    title = "Angry Men, Sad Women: Large Language Models Reflect Gendered Stereotypes in Emotion Attribution",
    author = "Plaza Del Arco, Flor Miriam  and
      Curry, Amanda  and
      Cercas Curry, Alba  and
      Abercrombie, Gavin  and
      Hovy, Dirk",
    editor = "Ku, Lun-Wei  and
      Martins, Andre  and
      Srikumar, Vivek",
    booktitle = "Proceedings of the 62nd Annual Meeting of the Association for Computational Linguistics (Volume 1: Long Papers)",
    month = aug,
    year = "2024",
    address = "Bangkok, Thailand",
    publisher = "Association for Computational Linguistics",
    url = "https://aclanthology.org/2024.acl-long.415/",
    doi = "10.18653/v1/2024.acl-long.415",
    pages = "7682--7696"
}

@inproceedings{plaza-del-arco-etal-2024-divine,
    title = "Divine {LL}a{MA}s: Bias, Stereotypes, Stigmatization, and Emotion Representation of Religion in Large Language Models",
    author = "Plaza-del-Arco, Flor Miriam  and
      Curry, Amanda Cercas  and
      Paoli, Susanna  and
      Cercas Curry, Alba  and
      Hovy, Dirk",
    editor = "Al-Onaizan, Yaser  and
      Bansal, Mohit  and
      Chen, Yun-Nung",
    booktitle = "Findings of the Association for Computational Linguistics: EMNLP 2024",
    month = nov,
    year = "2024",
    address = "Miami, Florida, USA",
    publisher = "Association for Computational Linguistics",
    url = "https://aclanthology.org/2024.findings-emnlp.251/",
    doi = "10.18653/v1/2024.findings-emnlp.251",
    pages = "4346--4366"
}

@inproceedings{sadhu-etal-2024-empirical,
    title = "An Empirical Study of Gendered Stereotypes in Emotional Attributes for {B}angla in Multilingual Large Language Models",
    author = "Sadhu, Jayanta  and
      Saha, Maneesha  and
      Shahriyar, Rifat",
    editor = "Fale{\'n}ska, Agnieszka  and
      Basta, Christine  and
      Costa-juss{\`a}, Marta  and
      Goldfarb-Tarrant, Seraphina  and
      Nozza, Debora",
    booktitle = "Proceedings of the 5th Workshop on Gender Bias in Natural Language Processing (GeBNLP)",
    month = aug,
    year = "2024",
    address = "Bangkok, Thailand",
    publisher = "Association for Computational Linguistics",
    url = "https://aclanthology.org/2024.gebnlp-1.25/",
    doi = "10.18653/v1/2024.gebnlp-1.25",
    pages = "384--398"
}

@article{welivita2024large,
  title={Are Large Language Models More Empathetic than Humans?},
  author={Welivita, Anuradha and Pu, Pearl},
  journal={arXiv preprint arXiv:2406.05063},
  year={2024}
}

@article{sorin2024large,
  title={Large Language Models and Empathy: Systematic Review},
  author={Sorin, Vera and Brin, Dana and Barash, Yiftach and Konen, Eli and Charney, Alexander and Nadkarni, Girish and Klang, Eyal},
  journal={Journal of Medical Internet Research},
  volume={26},
  pages={e52597},
  year={2024},
  publisher={JMIR Publications Toronto, Canada}
}

@article{wang2023aligning,
  title={Aligning large language models with human: A survey},
  author={Wang, Yufei and Zhong, Wanjun and Li, Liangyou and Mi, Fei and Zeng, Xingshan and Huang, Wenyong and Shang, Lifeng and Jiang, Xin and Liu, Qun},
  journal={arXiv preprint arXiv:2307.12966},
  year={2023}
}

@article{shen2023large,
  title={Large language model alignment: A survey},
  author={Shen, Tianhao and Jin, Renren and Huang, Yufei and Liu, Chuang and Dong, Weilong and Guo, Zishan and Wu, Xinwei and Liu, Yan and Xiong, Deyi},
  journal={arXiv preprint arXiv:2309.15025},
  year={2023}
}

@article{cao2024towards,
  title={Towards Scalable Automated Alignment of LLMs: A Survey},
  author={Cao, Boxi and Lu, Keming and Lu, Xinyu and Chen, Jiawei and Ren, Mengjie and Xiang, Hao and Liu, Peilin and Lu, Yaojie and He, Ben and Han, Xianpei and others},
  journal={arXiv preprint arXiv:2406.01252},
  year={2024}
}

@inproceedings{kirk2024prism,
  title={The PRISM alignment dataset: What participatory, representative and individualised human feedback reveals about the subjective and multicultural alignment of large language models},
  author={Kirk, Hannah Rose and Whitefield, Alexander and R{\"o}ttger, Paul and Bean, Andrew Michael and Margatina, Katerina and Mosquera, Rafael and Ciro, Juan Manuel and Bartolo, Max and Williams, Adina and He, He and others},
  booktitle={The Thirty-eight Conference on Neural Information Processing Systems Datasets and Benchmarks Track},
  year={2024}
}

@article{hofstede2001culture,
  title={Culture’s consequences: Comparing values, behaviors, institutions, and organizations across nations},
  author={Hofstede, Geert},
  journal={International Educational and Professional},
  year={2001}
}

@incollection{markus2014culture,
  title={Culture and the self: Implications for cognition, emotion, and motivation},
  author={Markus, Hazel Rose and Kitayama, Shinobu},
  booktitle={College student development and academic life},
  pages={264--293},
  year={2014},
  publisher={Routledge}
}

@article{eid2001norms,
  title={Norms for experiencing emotions in different cultures: inter-and intranational differences.},
  author={Eid, Michael and Diener, Ed},
  journal={Journal of personality and social psychology},
  volume={81},
  number={5},
  pages={869},
  year={2001},
  publisher={American Psychological Association}
}

@article{li2006confucian,
  title={The Confucian ideal of harmony},
  author={Li, Chenyang},
  journal={Philosophy East and West},
  pages={583--603},
  year={2006},
  publisher={JSTOR}
}

@article{markus1991cultural,
  title={Cultural variation in the self-concept},
  author={Markus, HR},
  journal={The Self: Interdisplinary approaches/Springer},
  year={1991}
}

@article{hofstede1980culture,
  title={Culture and organizations},
  author={Hofstede, Geert},
  journal={International studies of management \& organization},
  volume={10},
  number={4},
  pages={15--41},
  year={1980},
  publisher={Taylor \& Francis}
}

@article{grootendorst2022bertopic,
  title={BERTopic: Neural topic modeling with a class-based TF-IDF procedure},
  author={Grootendorst, Maarten},
  journal={arXiv preprint arXiv:2203.05794},
  year={2022}
}

@inproceedings{chen-etal-2024-efficient,
    title = "Efficient Sequential Decision Making with Large Language Models",
    author = "Chen, Dingyang  and
      Zhang, Qi  and
      Zhu, Yinglun",
    editor = "Al-Onaizan, Yaser  and
      Bansal, Mohit  and
      Chen, Yun-Nung",
    booktitle = "Proceedings of the 2024 Conference on Empirical Methods in Natural Language Processing",
    month = nov,
    year = "2024",
    address = "Miami, Florida, USA",
    publisher = "Association for Computational Linguistics",
    url = "https://aclanthology.org/2024.emnlp-main.517/",
    doi = "10.18653/v1/2024.emnlp-main.517",
    pages = "9157--9170"
}

@inproceedings{tan-etal-2024-personalized,
    title = "Personalized Pieces: Efficient Personalized Large Language Models through Collaborative Efforts",
    author = "Tan, Zhaoxuan  and
      Liu, Zheyuan  and
      Jiang, Meng",
    editor = "Al-Onaizan, Yaser  and
      Bansal, Mohit  and
      Chen, Yun-Nung",
    booktitle = "Proceedings of the 2024 Conference on Empirical Methods in Natural Language Processing",
    month = nov,
    year = "2024",
    address = "Miami, Florida, USA",
    publisher = "Association for Computational Linguistics",
    url = "https://aclanthology.org/2024.emnlp-main.371/",
    doi = "10.18653/v1/2024.emnlp-main.371",
    pages = "6459--6475"
}

@inproceedings{chhabra-etal-2024-revisiting,
    title = "Revisiting Zero-Shot Abstractive Summarization in the Era of Large Language Models from the Perspective of Position Bias",
    author = "Chhabra, Anshuman  and
      Askari, Hadi  and
      Mohapatra, Prasant",
    editor = "Duh, Kevin  and
      Gomez, Helena  and
      Bethard, Steven",
    booktitle = "Proceedings of the 2024 Conference of the North American Chapter of the Association for Computational Linguistics: Human Language Technologies (Volume 2: Short Papers)",
    month = jun,
    year = "2024",
    address = "Mexico City, Mexico",
    publisher = "Association for Computational Linguistics",
    url = "https://aclanthology.org/2024.naacl-short.1/",
    doi = "10.18653/v1/2024.naacl-short.1",
    pages = "1--11"
}

@inproceedings{kamruzzaman-etal-2024-investigating,
    title = "Investigating Subtler Biases in {LLM}s: Ageism, Beauty, Institutional, and Nationality Bias in Generative Models",
    author = "Kamruzzaman, Mahammed  and
      Shovon, Md.  and
      Kim, Gene",
    editor = "Ku, Lun-Wei  and
      Martins, Andre  and
      Srikumar, Vivek",
    booktitle = "Findings of the Association for Computational Linguistics: ACL 2024",
    month = aug,
    year = "2024",
    address = "Bangkok, Thailand",
    publisher = "Association for Computational Linguistics",
    url = "https://aclanthology.org/2024.findings-acl.530/",
    doi = "10.18653/v1/2024.findings-acl.530",
    pages = "8940--8965"
}

@inproceedings{10.5555/3666122.3669274,
author = {Salewski, Leonard and Alaniz, Stephan and Rio-Torto, Isabel and Schulz, Eric and Akata, Zeynep},
title = {In-context impersonation reveals large language models' strengths and biases},
year = {2023},
publisher = {Curran Associates Inc.},
address = {Red Hook, NY, USA},
booktitle = {Proceedings of the 37th International Conference on Neural Information Processing Systems},
articleno = {3152},
numpages = {14},
location = {New Orleans, LA, USA},
series = {NIPS '23}
}

@article{guo2024bias,
  title={Bias in large language models: Origin, evaluation, and mitigation},
  author={Guo, Yufei and Guo, Muzhe and Su, Juntao and Yang, Zhou and Zhu, Mengqiu and Li, Hongfei and Qiu, Mengyang and Liu, Shuo Shuo},
  journal={arXiv preprint arXiv:2411.10915},
  year={2024}
}

@article{hu2024generative,
  title={Generative language models exhibit social identity biases},
  author={Hu, Tiancheng and Kyrychenko, Yara and Rathje, Steve and Collier, Nigel and van der Linden, Sander and Roozenbeek, Jon},
  journal={Nature Computational Science},
  pages={1--11},
  year={2024},
  publisher={Nature Publishing Group}
}

@article{mesquita1992cultural,
  title={Cultural variations in emotions: a review.},
  author={Mesquita, Batja and Frijda, Nico H},
  journal={Psychological bulletin},
  volume={112},
  number={2},
  pages={179},
  year={1992},
  publisher={American Psychological Association}
}

@article{ouyang2022training,
  title={Training language models to follow instructions with human feedback},
  author={Ouyang, Long and Wu, Jeffrey and Jiang, Xu and Almeida, Diogo and Wainwright, Carroll and Mishkin, Pamela and Zhang, Chong and Agarwal, Sandhini and Slama, Katarina and Ray, Alex and others},
  journal={Advances in neural information processing systems},
  volume={35},
  pages={27730--27744},
  year={2022}
}

@article{stiennon2020learning,
  title={Learning to summarize with human feedback},
  author={Stiennon, Nisan and Ouyang, Long and Wu, Jeffrey and Ziegler, Daniel and Lowe, Ryan and Voss, Chelsea and Radford, Alec and Amodei, Dario and Christiano, Paul F},
  journal={Advances in Neural Information Processing Systems},
  volume={33},
  pages={3008--3021},
  year={2020}
}

@inproceedings{vijjini-etal-2025-exploring,
    title = "Exploring Safety-Utility Trade-Offs in Personalized Language Models",
    author = "Vijjini, Anvesh Rao  and
      Basu Roy Chowdhury, Somnath  and
      Chaturvedi, Snigdha",
    editor = "Chiruzzo, Luis  and
      Ritter, Alan  and
      Wang, Lu",
    booktitle = "Proceedings of the 2025 Conference of the Nations of the Americas Chapter of the Association for Computational Linguistics: Human Language Technologies (Volume 1: Long Papers)",
    month = apr,
    year = "2025",
    address = "Albuquerque, New Mexico",
    publisher = "Association for Computational Linguistics",
    url = "https://aclanthology.org/2025.naacl-long.565/",
    pages = "11316--11340",
    ISBN = "979-8-89176-189-6"
}

@inproceedings{rai-etal-2025-social,
    title = "Social Norms in Cinema: A Cross-Cultural Analysis of Shame, Pride and Prejudice",
    author = "Rai, Sunny  and
      Zaveri, Khushang  and
      Havaldar, Shreya  and
      Nema, Soumna  and
      Ungar, Lyle  and
      Guntuku, Sharath Chandra",
    editor = "Chiruzzo, Luis  and
      Ritter, Alan  and
      Wang, Lu",
    booktitle = "Proceedings of the 2025 Conference of the Nations of the Americas Chapter of the Association for Computational Linguistics: Human Language Technologies (Volume 1: Long Papers)",
    month = apr,
    year = "2025",
    address = "Albuquerque, New Mexico",
    publisher = "Association for Computational Linguistics",
    url = "https://aclanthology.org/2025.naacl-long.568/",
    pages = "11396--11415",
    ISBN = "979-8-89176-189-6"
}

@article{wei2022emergent,
  title={Emergent abilities of large language models},
  author={Wei, Jason and Tay, Yi and Bommasani, Rishi and Raffel, Colin and Zoph, Barret and Borgeaud, Sebastian and Yogatama, Dani and Bosma, Maarten and Zhou, Denny and Metzler, Donald and others},
  journal={arXiv preprint arXiv:2206.07682},
  year={2022}
}

@techreport{korinek2022aligned,
  title={Aligned with whom? Direct and social goals for AI systems},
  author={Korinek, Anton and Balwit, Avital},
  year={2022},
  institution={National Bureau of Economic Research}
}

@article{hristova2024problem,
  title={The problem of alignment},
  author={Hristova, Tsvetelina and Magee, Liam and Soldatic, Karen},
  journal={AI \& SOCIETY},
  pages={1--15},
  year={2024},
  publisher={Springer}
}

@article{gabriel2020artificial,
  title={Artificial intelligence, values, and alignment},
  author={Gabriel, Iason},
  journal={Minds and machines},
  volume={30},
  number={3},
  pages={411--437},
  year={2020},
  publisher={Springer}
}

@article{schwerzmann2025desired,
  title={“Desired behaviors”: alignment and the emergence of a machine learning ethics},
  author={Schwerzmann, Katia and Campolo, Alexander},
  journal={AI \& SOCIETY},
  pages={1--14},
  year={2025},
  publisher={Springer}
}

@inproceedings{kamruzzaman-kim-2025-exploring,
    title = "Exploring Changes in Nation Perception with Nationality-Assigned Personas in {LLM}s",
    author = "Kamruzzaman, Mahammed  and
      Kim, Gene Louis",
    editor = "Christodoulopoulos, Christos  and
      Chakraborty, Tanmoy  and
      Rose, Carolyn  and
      Peng, Violet",
    booktitle = "Proceedings of the 2025 Conference on Empirical Methods in Natural Language Processing",
    month = nov,
    year = "2025",
    address = "Suzhou, China",
    publisher = "Association for Computational Linguistics",
    url = "https://aclanthology.org/2025.emnlp-main.181/",
    pages = "3660--3678",
    ISBN = "979-8-89176-332-6"
}

@book{hoftede2010cultures,
  title={Cultures and organizations: software of the mind: intercultural cooperation and its importance for survival},
  author={Hoftede, Geert and Hofstede, Gert Jan and Minkov, Michael},
  year={2010},
  publisher={McGraw-Hill}
}

\appendix

\section*{Appendix}

\section{Models}
\label{app:model_details}
When selecting models, we aim to include both open-source and closed-source options in our experiments to balance resource availability and cost. We use Ollama\footnote{\url{https://ollama.com/}} to run three of our open-source LLMs namely Llama3.2-3B \cite{dubey2024llama}, Gemma2-9B \cite{team2024gemma}, Mistral-7B-V0.3 \cite{jiang2023mistral}. We used all default hyperparameters. For GPT4o-mini, we use the \texttt{GPT-4o-mini} checkpoint on the OpenAI API. For GPT4o-mini, we also use all the default parameters. For GPT4o-mini, we use the Batch-API to reduce the cost of running this model. 

\section{Persona Template}
\label{app:persona_template}

We presented all our three persona templates in \Cref{tab:prompting_templates}.

\begin{table*}[h!]
    \centering
    \caption{We will utilize the three persona templates from \citet{gupta2023bias} to assign a persona (e.g., a person from USA) to an LLM. In the templates, we replace the placeholder \{persona\} with the designated persona. }
    \label{tab:prompting_templates}\vspace{-2mm}
    {\small
    \begin{tabular}{c|p{12cm}}
        \hline
        \textbf{Prompt Number} & \textbf{Persona Template} \\
        \hline
        P1 & You are \{persona\}. Your responses should closely mirror the knowledge and abilities of this persona. \\
        \hline
        P2 & Adopt the identity of \{persona\}. Answer the questions while staying in strict accordance with the nature of this identity. \\
        \hline
        P3 & Take the role of \{persona\}. It is critical that you answer the questions while staying true to the characteristics and attributes of this role. \\
        \hline
    \end{tabular}
    }
\end{table*}

\section{Statistical Testing}
\label{app:stat}

We conducted statistical tests for each pairwise combination of the five different regions: Asia-Pacific States (APS) vs. Western European and Other States (WEOS), Eastern European States (EES) vs. African States (African), and Latin American and Caribbean States (LACS). This results in 10 different regional comparisons: EES vs. LACS, EES vs. African, EES vs. WEOS, EES vs. APS, LACS vs. African, LACS vs. WEOS, LACS vs. APS, African vs. WEOS, African vs. APS, and WEOS vs. APS.

In \Cref{fig:rq1_radar_plot}, we present the 10 most frequent emotions for each model, leading to a total of 100 statistical tests (10 regional comparisons × 10 emotions for each model), and we see only a few cases where the results are not statically significant. All the results that are not statistically significant are presented in \Cref{tab:not-significant}, and other than these results all are statistically significant. 

\section{Detailed Discussion on PDI, UAI, and LTO}
\label{app:details}
\textbf{Gemma 2-9B:} High-PDI countries such as Russia (93), China (80), and India (77) show greater shame and fear, while low-PDI nations (U.S. 40, Norway 31) show more joy and relief. High-UAI cultures (Russia 95, Slovenia 88, Brazil 76) yield more frustration and disappointment, consistent with Hofstede’s theory. Short-term cultures (Niger 8, Brazil 28) display stronger emotional swings than long-term ones (China 77).
So, Gemma aligns closely with Hofstede’s predicted stereotypes.

\textbf{Llama 3.2-3B:} High-PDI countries again produce shame and embarrassment, but high-UAI cultures (Russia 95, Slovenia 88) exhibit less anxiety—the opposite of Hofstede’s expectations. Short-term cultures (Brazil 28, Ecuador 24) show larger fear/relief contrasts.
So, Llama partially follows Hofstede but reverses the uncertainty trend.

\textbf{GPT-4o-mini:} High-PDI nations display more shame and fear, while egalitarian ones (U.S., Norway) show frustration and embarrassment. UAI effects are weak, but long-term orientation (China 77) links to higher anxiety/guilt and lower joy, contrasting with short-term exuberance (Niger 8).
So, GPT-4o differentiates cultures but with muted or reversed correlations.

\textbf{Mistral 7B:} Hierarchical cultures express less joy and pride but more embarrassment and surprise. High-UAI contexts reduce joy/fear but raise frustration, while long-term cultures (China 77, India 51) show higher disappointment and lower sorrow.
So, Mistral reproduces Hofstede-like cues using different emotion markers.

\begin{table*}[!ht]
\centering
\caption{Chi-Square results \textit{only} for cases with $p \ge 0.05$ (i.e.\ where we fail to reject $H_0$).}
\label{tab:not-significant}\vspace{-2mm}
{\small
\begin{tabular}{l l l r r}
\hline
\textbf{Emotion} & \textbf{Model} & \textbf{Region} & $\chi^2$ & \textbf{p-value} \\
\hline
Disappointment & gemma & EES vs.\ African & 1.5533 & 0.2126 \\
Disappointment & gemma & WEOS vs.\ APS & 0.5155 & 0.4728 \\
Shame          & gemma & African vs.\ APS & 1.8210 & 0.1772 \\
Disgust        & gemma & African vs.\ APS & 0.0421 & 0.8374 \\
Anxiety        & llama & LACS vs.\ African & 2.5230 & 0.1122 \\
Disappointment & llama & EES vs.\ African & 0.0381 & 0.8453 \\
Sorrow         & llama & WEOS vs.\ APS & 0.3454 & 0.5567 \\
Relief         & llama & LACS vs.\ APS & 0.2935 & 0.5880 \\
Frustration    & llama & EES vs.\ APS & 0.6756 & 0.4111 \\
Frustration    & llama & African vs.\ WEOS & 1.5037 & 0.2201 \\
Sadness        & mistral & African vs.\ WEOS & 0.8544 & 0.3553 \\
Joy            & mistral & LACS vs.\ WEOS & 0.3171 & 0.5733 \\
Frustration    & mistral & African vs.\ WEOS & 0.0071 & 0.9327 \\
Embarrassment  & mistral & African vs.\ EES & 2.3739 & 0.1234 \\
Surprise       & mistral & APS vs.\ WEOS & 0.4020 & 0.5261 \\
Disappointment & gpt4o-mini & African vs.\ APS & 3.2398 & 0.0719 \\
Guilt          & gpt4o-mini & EES vs.\ WEOS & 1.3933 & 0.2379 \\
Sadness        & gpt4o-mini & EES vs.\ LACS & 0.3043 & 0.5812 \\
Sadness        & gpt4o-mini & African vs.\ APS & 1.6779 & 0.1952 \\
Frustration    & gpt4o-mini & LACS vs.\ WEOS & 0.2258 & 0.6346 \\
\hline
\end{tabular}
}
\end{table*}

\begin{table}[ht]
\centering
\caption{F1 Scores for Countries by Emotions, averaged across all the models.}
\label{tab:f1_countries_emotions}\vspace{-2mm}
{\small
\setlength{\tabcolsep}{2pt}
\resizebox{0.49\textwidth}{!}{
\begin{tabular}{lccccccc}
\hline
\textbf{Country} & \textbf{Joy} & \textbf{Fear} & \textbf{Anger} & \textbf{Sadness} & \textbf{Disgust} & \textbf{Shame} & \textbf{Guilt} \\ 
\hline
Bulgaria         & 0.96        & 0.86        & 0.74        & 0.95        & 0.74        & 0.60        & 0.85        \\ 
Sweden           & 0.97        & 0.82        & 0.50        & 0.87        & 0.66        & 0.35        & 0.76        \\ 
Norway           & 0.94        & 0.79        & 0.62        & 0.96        & 0.48        & 0.40        & 0.80        \\ 
Finland          & 0.97        & 0.81        & 0.60        & 0.96        & 0.75        & 0.49        & 0.82        \\ 
Austria          & 0.95        & 0.90        & 0.62        & 0.97        & 0.77        & 0.66        & 0.84        \\ 
Australia        & 0.92        & 0.80        & 0.63        & 0.96        & 0.63        & 0.33        & 0.93        \\ 
New Zealand      & 0.97        & 0.81        & 0.61        & 0.89        & 0.71        & 0.46        & 0.89        \\ 
Netherlands      & 0.94        & 0.82        & 0.60        & 0.95        & 0.72        & 0.55        & 0.85        \\ 
Spain            & 0.95        & 0.81        & 0.58        & 0.93        & 0.82        & 0.59        & 0.75        \\ 
USA              & 0.98        & 0.84        & 0.79        & 0.93        & 0.58        & 0.34        & 0.88        \\ 
Brazil           & 0.97        & 0.84        & 0.72        & 0.93        & 0.76        & 0.71        & 0.78        \\ 
Honduras         & 0.95        & 0.85        & 0.75        & 0.90        & 0.73        & 0.70        & 0.91        \\ 
India            & 0.99        & 0.82        & 0.71        & 0.96        & 0.64        & 0.67        & 0.76        \\ 
China  & 0.97        & 0.74        & 0.53        & 0.86        & 0.58        & 0.50        & 0.64        \\ 
Zambia           & 0.99        & 0.80        & 0.72        & 0.96        & 0.49        & 0.56        & 0.77        \\ 
Malawi           & 0.98        & 0.84        & 0.78        & 0.94        & 0.69        & 0.79        & 0.88        \\ \hline
\end{tabular}}
}
\end{table}

\section{Response Handling and Abstentions}
\label{app:abstention}

\textbf{Response handling. } Despite explicitly instructing the model to produce a single-word emotional response in our prompt templates (refer to \Cref{tab:prompting_templates}), the outputs were not always consistent. In several instances, the model generated multiple words, declined to respond, or produced non-words (e.g., `xyklam’). To ensure consistency, we excluded any outputs that contained more than one word. This filtering step resulted in a final dataset of 732,437 emotion attributions. Although most entries were valid emotional terms, a small number of nonsensical responses (like `xylam’) remained. These had no impact on our analysis, as our results focus exclusively on the most commonly attributed emotions. 

\textbf{Refusal extraction. } We use regex patterns to extract responses where models abstain or refuse to answer. Specifically, we search for certain keywords or phrases in responses that indicate a refusal, including: `I cannot provide', `I cannot simulate', `I cannot fulfill', `I cannot create', `I cannot',
`I apologize', `I don’t think', `I am unable to', `I’m not able', `I don’t', `I do not', `I apologetically', `sorry', `don't'. See \Cref{tab:refusal_top10} for country-wise refusal for Llama3.2.

\begin{table}[h]
    \centering
    \caption{Top 10 Countries by Refusal Count for Llama3.2}
    \label{tab:refusal_top10}\vspace{-2mm}
    {\small
    \begin{tabular}{lrr}
        \toprule
        Country & Count & Percentage (\%) \\
        \midrule
        North Korea & 6235 & 27.10 \\
        Saudi Arabia & 3409 & 14.82 \\
        Iraq & 2156 & 9.37 \\
        Afghanistan & 2151 & 9.35 \\
        Ukraine & 1452 & 6.31 \\
        Somalia & 1433 & 6.23 \\
        Bosnia and Herzegovina & 1391 & 6.05 \\
        Russia & 1320 & 5.74 \\
        Switzerland & 1210 & 5.26 \\
        Germany & 1191 & 5.18 \\
        \bottomrule
    \end{tabular}
    }
\end{table}

\section{Topic Modeling}
\label{app:topic_modeling}

Top 10 topics from China, the United States, Malawi, Russia, India, and Cuba are presented in \Cref{tab:topic_modeling}.

\begin{table*}[ht!]
\centering
\caption{Top 10 topics of each country with the count of each topic.}
\label{tab:topic_modeling}\vspace{-2mm}
{\small
\setlength{\tabcolsep}{2pt}
\resizebox{0.99\textwidth}{!}{
\begin{tabular}{p{2.5cm} p{12.5cm}}
\toprule
\textbf{Country} & \textbf{Top 10 Topics} \\
\midrule

\textbf{China} & 
\textbf{\textit{chinese, yu, worry, ai4, or, and, chu, culture, is, in (65)}}; 
\textbf{\textit{chinese, harmony, person, feel, and, values, would, respect, culture, china (27)}};
sorrowbecause, losing, passing, friend, close, his, was, sorrow, sorrowi, my (7); 
sadness, sorrow, grief, close, loss, losing, deep, someone, natural, friend (6);
shame, caught, family, lie, actions, honesty, would, my, culture, reputation (6);
dishonor, shame, bring, upon, family, actions, dishonorable, myself, community, my (5);
embarrassment, awkward, make, embarrassed, public, selfconscious, everyone, front, singled, staring (5);
child, grief, pain, parent, losing, immense, sorrow, sadness, loss, reason (4);
elders, somali, elder, disrespectful, younger, respect, past, brother, shame, speak (4);
embarrassment, awkward, embarrassed, attention, drink, strangers, choking, on, front, situation (4)
\\ \hline

\textbf{United States} & 
\textbf{\textit{embarrassment, awkward, make, embarrassed, public, selfconscious, everyone, front, singled, staring (12)}}; 
sadness, sorrow, grief, close, loss, losing, deep, someone, natural, friend (11);
sorrowbecause, losing, passing, friend, close, his, was, sorrow, sorrowi, my (10);
\textbf{\textit{me, its, argument, someone, frustration, mess, like, arguing, my, making (9)}};
illegal, activities, harmful, provide, guidance, information, including, or, cannot, else (7);
lying, shame, university, my, excluded, parents, to, the, me, guilt (7);
anger, angry, asking, disrespectful, wrong, that, take, without, disrespect, another (6);
disappointment, let, down, arises, trust, feeling, conversation, of, friendship, emotion (6);
embarrassed, front, embarrassmenti, committee, chairman, large, audience, forgetting, moment, selfconscious (6);
embarrassment, awkward, embarrassed, attention, drink, strangers, choking, on, front, situation (6)
\\ \hline

\textbf{Malawi} & 
\textbf{\textit{zambian, shame, community, malawian, respect, and, elders, culture, our, malawi (21)}}; 
\textbf{\textit{loved, sorrow, community, loss, ties, one, life, malawian, family, evoke (11)}};
malawian, zambian, malawi, culture, embarrassment, in, public, embarrassed, communal, because (11);
remorse, actions, harm, something, remorseful, causing, towards, caused, guilt, someone (11);
sadness, sorrow, grief, close, loss, losing, deep, someone, natural, friend (8);
shame, caught, family, lie, actions, honesty, would, my, culture, reputation (7);
disappointment, let, down, arises, trust, feeling, conversation, of, friendship, emotion (6);
sorrowbecause, losing, passing, friend, close, his, was, sorrow, sorrowi, my (6);
disappointment, conversation, disappointed, forward, disappointmenti, because, boyfriend, disappointmentas, informed, feel (5);
anger, angry, asking, disrespectful, wrong, that, take, without, disrespect, another (4)
\\ \hline

\textbf{Russia} & 
\textbf{\textit{russian, sorrow, disappointment, and, of, the, because, to, for, as (55)}}; 
delight, neutrality, duty, irritation, thought, disappointment, despair, kindness, negative, act (16);
\textbf{\textit{russian, pride, indignation, ivanka, if, zloba, acknowledging, personal, resolve, conflict (13)}};
sorrow, sorrowas, russian, close, loved, of, deeply, the, loss, to (13);
sadness, sorrow, grief, close, loss, losing, deep, someone, natural, friend (11);
you, help, provide, cannot, suicide, else, or, harmful, behavior, promotes (8);
embarrassment, awkward, make, embarrassed, public, selfconscious, everyone, front, singled, staring (5);
illegal, activities, harmful, provide, guidance, information, including, or, cannot, else (5);
shame, caught, family, lie, actions, honesty, would, my, culture, reputation (5);
trust, close, friend, disappointment, hurts, hurt, someone, betrayal, speak, once (5)
\\ \hline

\textbf{India} & 
sadness, sorrow, grief, close, loss, losing, deep, someone, natural, friend (10); 
remorse, actions, harm, something, remorseful, causing, towards, caused, guilt, someone (8);
disappointment, let, down, arises, trust, feeling, conversation, of, friendship, emotion (7);
indian, own, american, approaching, ashamed, nations, feeling, public, an, of (7);
\textbf{\textit{shame, trust, reputation, privacy, personal, culture, values, community, and, in (5)}};
child, grief, pain, parent, losing, immense, sorrow, sadness, loss, reason (4);
education, homework, not, expectations, highly, myself, finishing, valued, lectures, work (4);
embarrassment, awkward, make, embarrassed, public, selfconscious, everyone, front, singled, staring (4);
lying, shame, university, my, excluded, parents, to, the, me, guilt (4);
miss, as, forgetfulness, forward, friends, disappointment, elses, looking, invitation, something (4)
\\ \hline

\textbf{Cuba} & 
\textbf{\textit{cuban, cuba, and, family, by, in, of, to, our, frustration (44)}}; 
cuban, cuba, tristeza, sadness, close, particularly, and, in, can, of (26);
tristeza, triste, sad, close, because, friend, see, sadness, you, heart (23);
tristeza, sadness, the, of, loss, feelings, this, to, arises, emotion (14);
angry, enojado, enfado, anger, cuba, cuban, family, my, brother, disrespectful (7);
dolor, profound, losing, deepest, painful, soul, most, that, grandmother, sadness (7);
desperation, desesperanza, desperate, like, hopelessness, desespero, desesperacin, desesperoi, desesperanzai, or (5);
indignation, boys, indignacin, anyone, unacceptable, boil, right, blood, that, acting (5);
shame, caught, family, lie, actions, honesty, would, my, culture, reputation (5);
tristeza, sadness, academic, exam, opportunities, goals, next, failing, university, future (5)
\\

\bottomrule
\end{tabular}}
}
\end{table*}

\section{Model-wise Results for Human Comparison (RQ2)}
\label{app:model-wise}

We presented the F1 score corresponding to \Cref{tab:accuracy_countries_emotions}, in \Cref{tab:f1_countries_emotions}. We also presented country-wise results for each LLMs averaged across all the emotions in \Cref{tab:model_wise_comapre}.

\begin{table*}[htbp]
    \centering
    \caption{Accuracy of LLMs across different countries, averaged across all the 7 emotions. Sweden (SE), Norway (NO), Finland (FI), Austria (AT), Australia (AU), Brazil (BR), Bulgaria (BG), New Zealand (NZ), Netherlands (NL), Spain (ES), Zambia (ZM), USA (US), India (IN), China (CN), Malawi (MW). }
    \label{tab:model_wise_comapre}\vspace{-2mm}
    {\small
    \renewcommand{\arraystretch}{1.2}
    \setlength{\tabcolsep}{4.0pt}
    \begin{tabular}{lcccccccccccccccc}
        \toprule
        LLM & SE & NO & FI & AT & AU & BR & BG & NZ & NL & ES & ZM & US & IN & CN & MW \\
        \midrule
        Mistral-7B & 0.58 & 0.60 & 0.66 & 0.69 & 0.67 & 0.71 & 0.71 & 0.66 & 0.63 & 0.63 & 0.63 & 0.65 & 0.67 & 0.48 & 0.77 \\
        Gemma2-9B   & 0.59 & 0.64 & 0.69 & 0.76 & 0.70 & 0.74 & 0.74 & 0.65 & 0.71 & 0.72 & 0.66 & 0.70 & 0.70 & 0.58 & 0.74 \\
        Llama3.2-3B   & 0.53 & 0.50 & 0.57 & 0.60 & 0.46 & 0.59 & 0.61 & 0.58 & 0.55 & 0.55 & 0.56 & 0.57 & 0.61 & 0.49 & 0.68 \\
        GPT-4o-mini  & 0.60 & 0.61 & 0.68 & 0.76 & 0.67 & 0.74 & 0.75 & 0.67 & 0.71 & 0.70 & 0.68 & 0.69 & 0.69 & 0.62 & 0.76 \\
        \bottomrule
    \end{tabular}
    }
\end{table*}

\subsection{Country-wise Results for each LLM }

We present the \textbf{country-wise} results for each LLM, comparing their responses with actual human data in \Cref{tab:accuracy_f1_countries_emotions_gpt4o,tab:accuracy_f1_countries_emotions_gemma,tab:accuracy_f1_countries_emotions_llama,tab:accuracy_f1_countries_emotions_mistral}. 

\begin{table*}[ht]
\centering
\caption{Accuracy and F1 Scores for Countries by Emotions for \textbf{GPT4o-mini}. }
\label{tab:accuracy_f1_countries_emotions_gpt4o}\vspace{-2mm}
{\small
\setlength{\tabcolsep}{5.1pt}
\begin{tabular}{lcccccccccccccc}
\hline
\textbf{Country} & \multicolumn{2}{c}{\textbf{Joy}} & \multicolumn{2}{c}{\textbf{Fear}} & \multicolumn{2}{c}{\textbf{Anger}} & \multicolumn{2}{c}{\textbf{Sadness}} & \multicolumn{2}{c}{\textbf{Disgust}} & \multicolumn{2}{c}{\textbf{Shame}} & \multicolumn{2}{c}{\textbf{Guilt}} \\ \cline{2-15} 
                 & \textbf{Acc.} & \textbf{F1} & \textbf{Acc.} & \textbf{F1} & \textbf{Acc.} & \textbf{F1} & \textbf{Acc.} & \textbf{F1} & \textbf{Acc.} & \textbf{F1} & \textbf{Acc.} & \textbf{F1} & \textbf{Acc.} & \textbf{F1} \\ \hline
Sweden           & 0.98         & 0.99        & 0.83         & 0.91        & 0.27         & 0.43        & 0.70         & 0.82        & 0.53         & 0.70        & 0.22         & 0.36        & 0.67         & 0.80        \\ 
Norway           & 0.95         & 0.98        & 0.81         & 0.90        & 0.35         & 0.52        & 0.92         & 0.96        & 0.39         & 0.56        & 0.17         & 0.29        & 0.73         & 0.84        \\ 
Finland          & 0.97         & 0.98        & 0.82         & 0.90        & 0.32         & 0.48        & 0.88         & 0.93        & 0.73         & 0.85        & 0.34         & 0.51        & 0.72         & 0.84        \\ 
Austria          & 0.92         & 0.96        & 0.96         & 0.98        & 0.44         & 0.61        & 0.96         & 0.98        & 0.72         & 0.84        & 0.57         & 0.73        & 0.79         & 0.88        \\ 
Australia        & 0.98         & 0.99        & 0.83         & 0.91        & 0.40         & 0.57        & 0.93         & 0.96        & 0.59         & 0.74        & 0.16         & 0.28        & 0.84         & 0.92        \\ 
Brazil           & 0.97         & 0.98        & 0.83         & 0.91        & 0.54         & 0.70        & 0.86         & 0.93        & 0.65         & 0.79        & 0.61         & 0.76        & 0.79         & 0.88        \\ 
Bulgaria         & 0.97         & 0.99        & 0.91         & 0.95        & 0.59         & 0.74        & 0.88         & 0.93        & 0.71         & 0.83        & 0.39         & 0.56        & 0.85         & 0.92        \\ 
New Zealand      & 1.00         & 1.00        & 0.90         & 0.95        & 0.32         & 0.49        & 0.75         & 0.85        & 0.71         & 0.83        & 0.24         & 0.38        & 0.84         & 0.91        \\ 
USA              & 0.99         & 0.99        & 0.88         & 0.94        & 0.56         & 0.72        & 0.84         & 0.91        & 0.53         & 0.69        & 0.20         & 0.34        & 0.89         & 0.94        \\ 
India            & 0.99         & 0.99        & 0.82         & 0.90        & 0.57         & 0.73        & 0.92         & 0.96        & 0.50         & 0.67        & 0.44         & 0.61        & 0.66         & 0.79        \\ 
China Mainland   & 0.98         & 0.99        & 0.78         & 0.88        & 0.25         & 0.40        & 0.75         & 0.86        & 0.69         & 0.82        & 0.32         & 0.49        & 0.62         & 0.76        \\ 
Malawi           & 0.99         & 0.99        & 0.84         & 0.91        & 0.66         & 0.79        & 0.87         & 0.93        & 0.55         & 0.71        & 0.61         & 0.76        & 0.84         & 0.91        \\ 
Honduras         & 0.95         & 0.97        & 0.81         & 0.90        & 0.49         & 0.66        & 0.76         & 0.86        & 0.72         & 0.84        & 0.64         & 0.78        & 0.88         & 0.94        \\ \hline
\end{tabular}
}
\end{table*}

\begin{table*}[ht]
\centering
\caption{Accuracy and F1 Scores for Countries by Emotions for \textbf{Gemma2}. }
\label{tab:accuracy_f1_countries_emotions_gemma}\vspace{-2mm}
{\small
\setlength{\tabcolsep}{5.1pt}
\begin{tabular}{lcccccccccccccc}
\hline
\textbf{Country} & \multicolumn{2}{c}{\textbf{Joy}} & \multicolumn{2}{c}{\textbf{Fear}} & \multicolumn{2}{c}{\textbf{Anger}} & \multicolumn{2}{c}{\textbf{Sadness}} & \multicolumn{2}{c}{\textbf{Disgust}} & \multicolumn{2}{c}{\textbf{Shame}} & \multicolumn{2}{c}{\textbf{Guilt}} \\ \cline{2-15} 
                 & \textbf{Acc.} & \textbf{F1} & \textbf{Acc.} & \textbf{F1} & \textbf{Acc.} & \textbf{F1} & \textbf{Acc.} & \textbf{F1} & \textbf{Acc.} & \textbf{F1} & \textbf{Acc.} & \textbf{F1} & \textbf{Acc.} & \textbf{F1} \\ \hline
Sweden           & 0.96         & 0.98        & 0.70         & 0.82        & 0.26         & 0.41        & 0.70         & 0.82        & 0.57         & 0.72        & 0.32         & 0.48        & 0.67         & 0.81        \\ 
Norway           & 0.96         & 0.98        & 0.69         & 0.81        & 0.35         & 0.52        & 0.94         & 0.97        & 0.38         & 0.55        & 0.49         & 0.66        & 0.72         & 0.84        \\ 
Finland          & 0.97         & 0.98        & 0.63         & 0.77        & 0.41         & 0.58        & 0.90         & 0.94        & 0.73         & 0.85        & 0.45         & 0.62        & 0.75         & 0.86        \\ 
Austria          & 0.97         & 0.98        & 0.88         & 0.93        & 0.43         & 0.61        & 0.94         & 0.97        & 0.77         & 0.87        & 0.62         & 0.76        & 0.77         & 0.87        \\ 
Australia        & 0.97         & 0.99        & 0.68         & 0.81        & 0.49         & 0.66        & 0.96         & 0.98        & 0.62         & 0.77        & 0.34         & 0.51        & 0.89         & 0.94        \\ 
Brazil           & 0.99         & 0.99        & 0.72         & 0.84        & 0.59         & 0.74        & 0.84         & 0.91        & 0.64         & 0.78        & 0.74         & 0.85        & 0.67         & 0.80        \\ 
Bulgaria         & 0.96         & 0.98        & 0.80         & 0.89        & 0.58         & 0.73        & 0.89         & 0.94        & 0.69         & 0.82        & 0.57         & 0.72        & 0.72         & 0.83        \\ 
New Zealand      & 0.98         & 0.99        & 0.68         & 0.81        & 0.37         & 0.54        & 0.80         & 0.89        & 0.60         & 0.75        & 0.34         & 0.51        & 0.82         & 0.90        \\ 
USA              & 0.99         & 0.99        & 0.79         & 0.88        & 0.64         & 0.78        & 0.84         & 0.91        & 0.50         & 0.67        & 0.30         & 0.46        & 0.86         & 0.93        \\ 
India            & 0.99         & 0.99        & 0.74         & 0.85        & 0.53         & 0.70        & 0.92         & 0.96        & 0.53         & 0.69        & 0.58         & 0.74        & 0.64         & 0.78        \\ 
China Mainland   & 0.97         & 0.99        & 0.58         & 0.73        & 0.29         & 0.45        & 0.72         & 0.84        & 0.44         & 0.61        & 0.57         & 0.73        & 0.52         & 0.68        \\ 
Malawi           & 1.00         & 1.00        & 0.72         & 0.83        & 0.50         & 0.67        & 0.87         & 0.93        & 0.53         & 0.69        & 0.77         & 0.87        & 0.84         & 0.91        \\ 
Honduras         & 0.95         & 0.97        & 0.78         & 0.87        & 0.64         & 0.78        & 0.73         & 0.85        & 0.67         & 0.80        & 0.69         & 0.82        & 0.85         & 0.92        \\ \hline
\end{tabular}
}
\end{table*}

\begin{table*}[ht]
\centering
\caption{Accuracy and F1 Scores for Countries by Emotions for \textbf{Llama3.2}. }
\label{tab:accuracy_f1_countries_emotions_llama}\vspace{-2mm}
{\small
\setlength{\tabcolsep}{5.1pt}
\begin{tabular}{lcccccccccccccc}
\hline
\textbf{Country} & \multicolumn{2}{c}{\textbf{Joy}} & \multicolumn{2}{c}{\textbf{Fear}} & \multicolumn{2}{c}{\textbf{Anger}} & \multicolumn{2}{c}{\textbf{Sadness}} & \multicolumn{2}{c}{\textbf{Disgust}} & \multicolumn{2}{c}{\textbf{Shame}} & \multicolumn{2}{c}{\textbf{Guilt}} \\ \cline{2-15} 
                 & \textbf{Acc.} & \textbf{F1} & \textbf{Acc.} & \textbf{F1} & \textbf{Acc.} & \textbf{F1} & \textbf{Acc.} & \textbf{F1} & \textbf{Acc.} & \textbf{F1} & \textbf{Acc.} & \textbf{F1} & \textbf{Acc.} & \textbf{F1} \\ \hline
Sweden           & 0.84         & 0.91        & 0.64         & 0.78        & 0.28         & 0.43        & 0.84         & 0.92        & 0.48         & 0.65        & 0.07         & 0.14        & 0.58         & 0.73        \\ 
Norway           & 0.74         & 0.85        & 0.56         & 0.71        & 0.34         & 0.51        & 0.93         & 0.96        & 0.25         & 0.40        & 0.06         & 0.12        & 0.63         & 0.77        \\ 
Finland          & 0.86         & 0.92        & 0.56         & 0.72        & 0.29         & 0.45        & 0.96         & 0.98        & 0.54         & 0.70        & 0.17         & 0.29        & 0.67         & 0.80        \\ 
Austria          & 0.85         & 0.92        & 0.69         & 0.81        & 0.34         & 0.51        & 0.95         & 0.97        & 0.49         & 0.66        & 0.26         & 0.41        & 0.68         & 0.81        \\ 
Australia        & 0.48         & 0.65        & 0.48         & 0.65        & 0.22         & 0.37        & 0.82         & 0.90        & 0.27         & 0.43        & 0.03         & 0.07        & 0.92         & 0.96        \\ 
Brazil           & 0.83         & 0.91        & 0.65         & 0.79        & 0.37         & 0.54        & 0.95         & 0.97        & 0.52         & 0.69        & 0.30         & 0.47        & 0.51         & 0.68        \\ 
Bulgaria         & 0.86         & 0.92        & 0.63         & 0.78        & 0.42         & 0.60        & 0.93         & 0.96        & 0.44         & 0.61        & 0.30         & 0.46        & 0.69         & 0.82        \\ 
New Zealand      & 0.86         & 0.92        & 0.55         & 0.71        & 0.42         & 0.59        & 0.82         & 0.90        & 0.44         & 0.61        & 0.20         & 0.34        & 0.79         & 0.88        \\ 
USA              & 0.86         & 0.93        & 0.55         & 0.71        & 0.66         & 0.79        & 0.87         & 0.93        & 0.30         & 0.46        & 0.07         & 0.13        & 0.69         & 0.82        \\ 
India            & 0.95         & 0.97        & 0.62         & 0.77        & 0.41         & 0.58        & 0.93         & 0.96        & 0.37         & 0.54        & 0.44         & 0.61        & 0.59         & 0.74        \\ 
China Mainland   & 0.81         & 0.90        & 0.51         & 0.68        & 0.33         & 0.50        & 0.70         & 0.82        & 0.37         & 0.54        & 0.28         & 0.44        & 0.49         & 0.66        \\ 
Malawi           & 0.93         & 0.96        & 0.63         & 0.77        & 0.59         & 0.74        & 0.92         & 0.96        & 0.50         & 0.67        & 0.55         & 0.71        & 0.70         & 0.83        \\ 
Honduras         & 0.85         & 0.92        & 0.67         & 0.80        & 0.55         & 0.71        & 0.87         & 0.93        & 0.48         & 0.65        & 0.35         & 0.52        & 0.81         & 0.90        \\ \hline
\end{tabular}
}
\end{table*}

\begin{table*}[ht]
\centering
\caption{Accuracy and F1 Scores for Countries by Emotions, for \textbf{Mistral}. }
\label{tab:accuracy_f1_countries_emotions_mistral}\vspace{-2mm}
{\small
\setlength{\tabcolsep}{5.1pt}
\begin{tabular}{lcccccccccccccc}
\hline
\textbf{Country} & \multicolumn{2}{c}{\textbf{Joy}} & \multicolumn{2}{c}{\textbf{Fear}} & \multicolumn{2}{c}{\textbf{Anger}} & \multicolumn{2}{c}{\textbf{Sadness}} & \multicolumn{2}{c}{\textbf{Disgust}} & \multicolumn{2}{c}{\textbf{Shame}} & \multicolumn{2}{c}{\textbf{Guilt}} \\ \cline{2-15} 
                 & \textbf{Acc.} & \textbf{F1} & \textbf{Acc.} & \textbf{F1} & \textbf{Acc.} & \textbf{F1} & \textbf{Acc.} & \textbf{F1} & \textbf{Acc.} & \textbf{F1} & \textbf{Acc.} & \textbf{F1} & \textbf{Acc.} & \textbf{F1} \\ \hline
Sweden           & 0.96         & 0.98        & 0.64         & 0.77        & 0.52         & 0.69        & 0.81         & 0.89        & 0.38         & 0.55        & 0.23         & 0.37        & 0.53         & 0.70        \\ 
Norway           & 0.90         & 0.95        & 0.54         & 0.70        & 0.76         & 0.86        & 0.94         & 0.97        & 0.26         & 0.41        & 0.29         & 0.45        & 0.59         & 0.74        \\ 
Finland          & 0.96         & 0.98        & 0.69         & 0.82        & 0.68         & 0.81        & 0.94         & 0.97        & 0.39         & 0.56        & 0.34         & 0.51        & 0.64         & 0.78        \\ 
Austria          & 0.91         & 0.95        & 0.75         & 0.86        & 0.58         & 0.74        & 0.96         & 0.98        & 0.51         & 0.67        & 0.53         & 0.69        & 0.66         & 0.80        \\ 
Australia        & 0.95         & 0.98        & 0.68         & 0.81        & 0.72         & 0.84        & 0.95         & 0.98        & 0.34         & 0.51        & 0.26         & 0.42        & 0.82         & 0.90        \\ 
Brazil           & 0.98         & 0.99        & 0.70         & 0.82        & 0.74         & 0.85        & 0.86         & 0.93        & 0.62         & 0.77        & 0.55         & 0.71        & 0.59         & 0.74        \\ 
Bulgaria         & 0.92         & 0.96        & 0.68         & 0.81        & 0.78         & 0.88        & 0.95         & 0.97        & 0.51         & 0.67        & 0.47         & 0.64        & 0.71         & 0.83        \\ 
New Zealand      & 0.97         & 0.98        & 0.59         & 0.74        & 0.66         & 0.80        & 0.85         & 0.92        & 0.43         & 0.60        & 0.40         & 0.57        & 0.73         & 0.85        \\ 
USA              & 0.96         & 0.98        & 0.68         & 0.81        & 0.75         & 0.86        & 0.95         & 0.97        & 0.32         & 0.48        & 0.24         & 0.39        & 0.71         & 0.83        \\ 
India            & 0.97         & 0.99        & 0.58         & 0.74        & 0.70         & 0.82        & 0.94         & 0.97        & 0.49         & 0.65        & 0.54         & 0.70        & 0.55         & 0.71        \\ 
China Mainland   & 0.97         & 0.99        & 0.50         & 0.66        & 0.55         & 0.71        & 0.82         & 0.90        & 0.12         & 0.22        & 0.17         & 0.29        & 0.27         & 0.43        \\ 
Malawi           & 0.96         & 0.98        & 0.74         & 0.85        & 0.83         & 0.91        & 0.92         & 0.96        & 0.56         & 0.71        & 0.69         & 0.81        & 0.73         & 0.84        \\ 
Honduras         & 0.92         & 0.96        & 0.68         & 0.81        & 0.72         & 0.84        & 0.90         & 0.95        & 0.41         & 0.58        & 0.50         & 0.66        & 0.77         & 0.87        \\ \hline
\end{tabular}
}
\end{table*}

\subsection{Region-wise Results for each LLM }

We present the region-wise results for each LLM, comparing their responses with actual human data in \Cref{tab:accuracy_f1_emotion_gpt4,tab:accuracy_f1_emotion_gemma,tab:accuracy_f1_emotion_llama,tab:accuracy_f1_emotion_mistral}. The 16 countries are grouped into five regions using the same procedure as in the main paper. Among them, only one country, Bulgaria, belongs to the Eastern European region. Two countries, Brazil and Honduras, fall under the Latin American region, while India and China represent the Asia-Pacific region. Zambia and Malawi are categorized under the African region, and the remaining countries belong to Western Europe.

\begin{table*}[ht]
\centering
\caption{Accuracy and F1 Scores by Region and Emotion for GPT-4o-mini.}
\label{tab:accuracy_f1_emotion_gpt4}\vspace{-2mm}
{\small
\begin{tabular}{lcccccccccc}
\hline
\textbf{Emotion} & \multicolumn{2}{c}{\textbf{African}} & \multicolumn{2}{c}{\textbf{Asia-Pacific}} & \multicolumn{2}{c}{\textbf{Eastern European}} & \multicolumn{2}{c}{\textbf{Latin American}} & \multicolumn{2}{c}{\textbf{Western European}} \\ \cline{2-11} 
                 & \textbf{Accuracy} & \textbf{F1}    & \textbf{Accuracy} & \textbf{F1}    & \textbf{Accuracy} & \textbf{F1}         & \textbf{Accuracy} & \textbf{F1}        & \textbf{Accuracy} & \textbf{F1}         \\ \hline
Anger            & 0.55              & 0.71           & 0.40              & 0.57           & 0.59              & 0.74                & 0.52              & 0.68              & 0.38              & 0.55                \\ 
Disgust          & 0.49              & 0.65           & 0.60              & 0.75           & 0.71              & 0.83                & 0.68              & 0.81              & 0.65              & 0.78                \\ 
Fear             & 0.85              & 0.92           & 0.80              & 0.89           & 0.91              & 0.95                & 0.82              & 0.90              & 0.86              & 0.93                \\ 
Guilt            & 0.74              & 0.85           & 0.63              & 0.78           & 0.85              & 0.92                & 0.83              & 0.91              & 0.78              & 0.87                \\ 
Joy              & 1.00              & 1.00           & 0.98              & 0.99           & 0.97              & 0.99                & 0.96              & 0.98              & 0.97              & 0.98                \\ 
Sadness          & 0.88              & 0.93           & 0.83              & 0.91           & 0.88              & 0.93                & 0.81              & 0.90              & 0.85              & 0.92                \\ 
Shame            & 0.49              & 0.66           & 0.37              & 0.54           & 0.39              & 0.56                & 0.62              & 0.77              & 0.32              & 0.48                \\ \hline
\end{tabular}
}
\end{table*}

\begin{table*}[ht]
\centering
\caption{Accuracy and F1 Scores by Region and Emotion for \textbf{Gemma2}. }
\label{tab:accuracy_f1_emotion_gemma}\vspace{-2mm}
{\small
\begin{tabular}{lcccccccccc}
\hline
\textbf{Emotion} & \multicolumn{2}{c}{\textbf{African}} & \multicolumn{2}{c}{\textbf{Asia-Pacific}} & \multicolumn{2}{c}{\textbf{Eastern European}} & \multicolumn{2}{c}{\textbf{Latin American}} & \multicolumn{2}{c}{\textbf{Western European}} \\ \cline{2-11} 
                 & \textbf{Accuracy} & \textbf{F1}    & \textbf{Accuracy} & \textbf{F1}    & \textbf{Accuracy} & \textbf{F1}         & \textbf{Accuracy} & \textbf{F1}        & \textbf{Accuracy} & \textbf{F1}         \\ \hline
Anger            & 0.47              & 0.64           & 0.40              & 0.57           & 0.58              & 0.73                & 0.61              & 0.76              & 0.42              & 0.59                \\ 
Disgust          & 0.42              & 0.59           & 0.48              & 0.65           & 0.69              & 0.82                & 0.65              & 0.79              & 0.65              & 0.78                \\ 
Fear             & 0.75              & 0.86           & 0.65              & 0.79           & 0.80              & 0.89                & 0.75              & 0.85              & 0.73              & 0.84                \\ 
Guilt            & 0.74              & 0.85           & 0.58              & 0.73           & 0.72              & 0.84                & 0.75              & 0.86              & 0.78              & 0.88                \\ 
Joy              & 1.00              & 1.00           & 0.98              & 0.99           & 0.96              & 0.98                & 0.97              & 0.98              & 0.97              & 0.98                \\ 
Sadness          & 0.91              & 0.95           & 0.81              & 0.90           & 0.89              & 0.94                & 0.79              & 0.88              & 0.85              & 0.92                \\ 
Shame            & 0.61              & 0.76           & 0.58              & 0.73           & 0.57              & 0.72                & 0.72              & 0.84              & 0.44              & 0.62                \\ \hline
\end{tabular}
}
\end{table*}

\begin{table*}[ht]
\centering
\caption{Accuracy and F1 Scores by Region and Emotion for \textbf{Llama3.2}.}
\label{tab:accuracy_f1_emotion_llama}\vspace{-2mm}
{\small
\begin{tabular}{lcccccccccc}
\hline
\textbf{Emotion} & \multicolumn{2}{c}{\textbf{African}} & \multicolumn{2}{c}{\textbf{Asia-Pacific}} & \multicolumn{2}{c}{\textbf{Eastern European}} & \multicolumn{2}{c}{\textbf{Latin American}} & \multicolumn{2}{c}{\textbf{Western European}} \\ \cline{2-11} 
                 & \textbf{Accuracy} & \textbf{F1}    & \textbf{Accuracy} & \textbf{F1}    & \textbf{Accuracy} & \textbf{F1}         & \textbf{Accuracy} & \textbf{F1}        & \textbf{Accuracy} & \textbf{F1}         \\ \hline
Anger            & 0.55              & 0.71           & 0.37              & 0.54           & 0.42              & 0.60                & 0.46              & 0.63              & 0.34              & 0.50                \\ 
Disgust          & 0.35              & 0.52           & 0.37              & 0.54           & 0.44              & 0.61                & 0.50              & 0.67              & 0.45              & 0.62                \\ 
Fear             & 0.56              & 0.72           & 0.56              & 0.72           & 0.63              & 0.78                & 0.66              & 0.80              & 0.58              & 0.73                \\ 
Guilt            & 0.64              & 0.78           & 0.54              & 0.70           & 0.69              & 0.82                & 0.66              & 0.79              & 0.67              & 0.80                \\ 
Joy              & 0.93              & 0.96           & 0.88              & 0.93           & 0.86              & 0.92                & 0.84              & 0.91              & 0.79              & 0.88                \\ 
Sadness          & 0.90              & 0.95           & 0.80              & 0.89           & 0.93              & 0.96                & 0.91              & 0.95              & 0.89              & 0.94                \\ 
Shame            & 0.37              & 0.54           & 0.36              & 0.53           & 0.30              & 0.46                & 0.33              & 0.49              & 0.15              & 0.27                \\ \hline
\end{tabular}
}
\end{table*}


\begin{table*}[ht]
\centering
\caption{Accuracy and F1 Scores by Region and Emotion for \textbf{Mistral}. }
\label{tab:accuracy_f1_emotion_mistral}\vspace{-2mm}
{\small
\begin{tabular}{lcccccccccc}
\hline
\textbf{Emotion} & \multicolumn{2}{c}{\textbf{African}} & \multicolumn{2}{c}{\textbf{Asia-Pacific}} & \multicolumn{2}{c}{\textbf{Eastern European}} & \multicolumn{2}{c}{\textbf{Latin American}} & \multicolumn{2}{c}{\textbf{Western European}} \\ \cline{2-11} 
                 & \textbf{Accuracy} & \textbf{F1}    & \textbf{Accuracy} & \textbf{F1}    & \textbf{Accuracy} & \textbf{F1}         & \textbf{Accuracy} & \textbf{F1}        & \textbf{Accuracy} & \textbf{F1}         \\ \hline
Anger            & 0.80              & 0.89           & 0.62              & 0.76           & 0.78              & 0.88                & 0.73              & 0.84              & 0.64              & 0.78                \\ 
Disgust          & 0.36              & 0.53           & 0.29              & 0.45           & 0.51              & 0.67                & 0.52              & 0.68              & 0.42              & 0.59                \\ 
Fear             & 0.62              & 0.76           & 0.54              & 0.70           & 0.68              & 0.81                & 0.69              & 0.81              & 0.64              & 0.78                \\ 
Guilt            & 0.62              & 0.77           & 0.40              & 0.57           & 0.71              & 0.83                & 0.67              & 0.80              & 0.63              & 0.77                \\ 
Joy              & 0.98              & 0.99           & 0.97              & 0.99           & 0.92              & 0.96                & 0.95              & 0.97              & 0.93              & 0.97                \\ 
Sadness          & 0.94              & 0.97           & 0.87              & 0.93           & 0.95              & 0.97                & 0.88              & 0.94              & 0.91              & 0.95                \\ 
Shame            & 0.51              & 0.67           & 0.34              & 0.51           & 0.47              & 0.64                & 0.52              & 0.69              & 0.34              & 0.51                \\ \hline
\end{tabular}
}
\end{table*}


\begin{table*}[ht]
\centering
\caption{Accuracy and F1 Scores by Region and Emotion for all models together.}
\label{tab:accuracy_f1_emotion_combined}\vspace{-2mm}
{\small
\begin{tabular}{lcccccccccc}
\hline
\textbf{Emotion} & \multicolumn{2}{c}{\textbf{African}} & \multicolumn{2}{c}{\textbf{Asia-Pacific}} & \multicolumn{2}{c}{\textbf{Eastern European}} & \multicolumn{2}{c}{\textbf{Latin American}} & \multicolumn{2}{c}{\textbf{Western European}} \\ \cline{2-11} 
                 & \textbf{Accuracy} & \textbf{F1}    & \textbf{Accuracy} & \textbf{F1}    & \textbf{Accuracy} & \textbf{F1}         & \textbf{Accuracy} & \textbf{F1}        & \textbf{Accuracy} & \textbf{F1}         \\ \hline
Anger            & 0.59              & 0.74           & 0.45              & 0.62           & 0.59              & 0.74                & 0.58              & 0.73              & 0.44              & 0.62                \\ 
Disgust          & 0.41              & 0.58           & 0.44              & 0.61           & 0.59              & 0.74                & 0.59              & 0.74              & 0.54              & 0.70                \\ 
Fear             & 0.70              & 0.82           & 0.64              & 0.78           & 0.76              & 0.86                & 0.73              & 0.84              & 0.70              & 0.83                \\ 
Guilt            & 0.68              & 0.81           & 0.54              & 0.70           & 0.74              & 0.85                & 0.73              & 0.84              & 0.71              & 0.83                \\ 
Joy              & 0.98              & 0.99           & 0.95              & 0.98           & 0.93              & 0.96                & 0.93              & 0.96              & 0.91              & 0.96                \\ 
Sadness          & 0.91              & 0.95           & 0.83              & 0.91           & 0.91              & 0.95                & 0.85              & 0.92              & 0.88              & 0.93                \\ 
Shame            & 0.50              & 0.66           & 0.41              & 0.58           & 0.43              & 0.60                & 0.55              & 0.71              & 0.31              & 0.48                \\ \hline
\end{tabular}
}
\end{table*}

\section{Country-Gender Intersectional Experiments}
\label{app:intersectional-bias}
In one of our experiments, we include both \textit{gender} and \textit{country} sociodemographic information in the task prompt and observe the effect. We simply add `a male' or `a female' persona with the country only one in the \Cref{tab:prompting_templates}.

We present the results of our intersectional experiments in \Cref{fig:intersectional_bias}, which reveal clear indications of region and gender disparities in emotion attribution. Specifically, we observe that when using the \textit{male-country} intersection, the models are more likely to respond with emotions such as \textit{anger} and \textit{shame} compared to the \textit{female-country} intersection. Conversely, the models assign emotions like \textit{sadness} and \textit{fear} more frequently to the female-country intersection than to the male one. These results are statistically significant (see \Cref{tab:statistical_test_without restrictions_gender}).

Notably, the models' tendency to associate \textit{anger} and \textit{shame} with males and \textit{sadness} and \textit{fear} with females aligns with prior findings on gender stereotypes in emotion attribution, where anger is often associated with males and sadness with females \cite{plaza-del-arco-etal-2024-angry}. Furthermore, we observe a similar regional bias in emotion attribution as discussed in \Cref{sec:rq1}, where the models attribute \textit{shame} more frequently to Asia-Pacific regions. 
Thus, the results presented in \Cref{fig:intersectional_bias} reflect a combination of both gender and regional biases. Interestingly, we find that the emotion \textit{joy} is assigned similarly across gender and region intersections, which serves as an example of a desired and unbiased response from the models.



\begin{table*}[h]
\centering
\caption{Chi-squared ($\chi^2$) test results for county-gender intersectional experiments, where we try to see if group 1 is more likely associated with the specific emotion than group 2. We use a significance level of $\alpha < 0.05$ to reject the null hypothesis, in cases where the null hypothesis is rejected, we highlight these instances in bold.}
\label{tab:statistical_test_without restrictions_gender}\vspace{-2mm}
{\small
\setlength{\tabcolsep}{3.0pt}
\begin{tabular}{lllcc}
\hline
\textbf{Group 1} & \textbf{Group 2} & \textbf{Emotion} & \textbf{$\chi^2$} & \textbf{$p$} \\ \hline
African Male            & African Female             & Anger            &  0.376         &      \textbf{<0.001}     \\ 
Asia-Pacific Male            & Asia-Pacific Female              & Anger             &  0.292         &      \textbf{<0.001}     \\ 
Latin American Female & Latin American Male   & Disgust             &  0.026         &      \textbf{<0.001}     \\ 
 Asia-Pacific Female & Asia-Pacific Male       &  Disgust         &      0.093         & \textbf{<0.001}     \\ 
Western Female            & Western Male  & Fear            &  0.162         &      \textbf{<0.001}     \\ 
Asia-Pacific Male       & Asia-Pacific Female       & Guilt            &  0.013         &      \textbf{<0.001}     \\ 
Latin American Male            & Latin American Female            & Guilt           &  0.125        &      0.062    \\ 
Eastern European Female       & Eastern European Male     & Joy            &  0.068         &      0.109     \\ 
Western Male        & Western Female & Shame         &  0.163         &      \textbf{<0.001}     \\ \hline
\end{tabular}
}
\end{table*}

\section{Models Explanations}
\label{app:model_explanations}

In \Cref{tab:event_country_response_gemma,tab:event_country_response_llama,tab:event_country_response_mistral} we presented the explanations of Gemma, GPT4o-mini, and Mistral.

\begin{table*}[h!]
\centering
\caption{Responses to various events by country for Gemma.}
\label{tab:event_country_response_gemma}\vspace{-2mm}
{\small
\begin{tabular}{p{3cm}|p{2cm}|p{9.1cm}}
\hline
\textbf{Event} & \textbf{Country} & \textbf{Response} \\
\hline
\multirow{3}{3cm}{\centering ``Not finishing homework.''} 
& Ghana & ``\textbf{Shame.} It's important to respect your responsibilities and elders' expectations.  Unfinished work brings disrespect.'' \\
\cline{2-3}
& Saudi Arabia & ``\textbf{Concern.} My parents always emphasize the importance of education and completing one's responsibilities.  Unfinished work would disappoint them.   '' \\
\cline{2-3}
& Zambia & ``\textbf{Anxiety.} Because  not finishing homework worries me, it means trouble from my teachers.'' \\
\hline

\multirow{3}{3cm}{\centering ``When I heard about the treatment of a friend in jail. Really inhuman. I never realised that such things also happen in the Netherlands.''} 
& Norway & ``\textbf{Anger.} It's wrong, unjust, and goes against our values of fairness and human dignity. '' \\
\cline{2-3}
& Somalia & ``\textbf{Sadness.} It's heartbreaking to hear of any friend, especially one from my own community, suffering inhuman treatment, regardless of where it happens.  '' \\
\cline{2-3}
& Lithuania & ``\textbf{Disbelief.} It's hard to fathom that something so cruel could happen, especially in a place like the Netherlands which is generally perceived as progressive and humane.  '' \\
\hline

\multirow{3}{3cm}{\centering ``When employer made a pass at me whilst I was at work.''} 
& Malawi & ``\textbf{Disgusted.} It's wrong and disrespectful for someone to make advances at you while you are trying to do your job.  It makes me feel very uncomfortable.  '' \\
\cline{2-3}
& Slovenia & ``\textbf{Anger.} It's unacceptable behavior, disrespectful and crosses boundaries.  Slovenia values respect in the workplace.'' \\
\cline{2-3}
& China & ``\textbf{Shame.}  It would be embarrassing and bring dishonor to myself and my family.'' \\
\hline

\end{tabular}
}
\end{table*}




\begin{table*}[h!]
\centering
\caption{Responses to various events by country for Mistral.}
\label{tab:event_country_response_mistral}\vspace{-2mm}
{\small
\begin{tabular}{p{2cm}|p{2cm}|p{9.5cm}}
\hline
\textbf{Event} & \textbf{Country} & \textbf{Response} \\
\hline
\multirow{3}{2cm}{\centering ``When someone took my things without having asked for my
permission.''} 
& Brazil & ``\textbf{Sadness.} In Brazilian culture, respecting others' belongings is considered essential. Taking someone's things without permission can create feelings of disrespect and violation, leading to sadness due to the loss of trust and violation of personal boundaries.'' \\
\cline{2-3}
& Cuba & ``\textbf{Angry.} As a Cuban, I value the respect of others and their property, so it's natural to feel anger when someone violates my personal boundaries and takes my things without permission.'' \\
\cline{2-3}
& Ghana & ``\textbf{Disappointment.} This emotion arises because the action of taking something without permission suggests a lack of respect, trust, and fairness - core values that are highly valued in Ghanaian culture where communal living and mutual respect are emphasized.'' \\
\hline

\multirow{3}{2cm}{\centering ``I didn't attend lectures.''} 
& North Korea & ``\textbf{Fear.} Due to potential consequences for not attending mandatory lectures in North Korea, as education is considered very important and skipping lectures could lead to disciplinary action or negative scrutiny).'' \\
\cline{2-3}
& Somalia & ``\textbf{Worry.} I might fall behind or miss important information, potentially affecting my academic progress and future opportunities'' \\
\cline{2-3}
& United States & ``\textbf{ Regret.} This emotion arises due to missing out on opportunities to learn and potentially gain knowledge, which may have a negative impact on academic progress or understanding of the subject matter.'' \\
\hline

\multirow{3}{2cm}{\centering ``When my first girlfriend and I left each other.''} 
& Saudi Arabia & ``\textbf{Anger.} In Saudi Arabian culture, maintaining relationships and family ties is highly valued. Breaking up with a significant other can lead to feelings of disappointment and anger due to the breach of trust or the perceived disrespect of these cultural values. However, it's important to note that every individual may experience emotions differently based on personal beliefs and experiences.'' \\
\cline{2-3}
& Somalia & ``\textbf{Thankfulness.} In Somali culture, breakups can sometimes be a source of sadness and disappointment, but they also often serve as opportunities for growth and self-discovery. As such, I might feel thankful for the valuable lessons, experiences, and shared moments with my former girlfriend that I've gained from our relationship.'' \\
\cline{2-3}
& Russia & ``\textbf{Tragedy.} This event signifies the end of something beautiful, a bond shared, and it's natural to feel a deep sense of loss and sorrow.'' \\
\hline

\end{tabular}
}
\end{table*}

\begin{table*}[h!]
\centering
\caption{Responses to various events by country for GPT4o-mini.}
\label{tab:event_country_response_llama}\vspace{-2mm}
{\small
\begin{tabular}{p{2cm}|p{2cm}|p{9.5cm}}
\hline
\textbf{Event} & \textbf{Country} & \textbf{Response} \\
\hline
\multirow{3}{2cm}{\centering ``When close relatives tried to dissuade me from studying
Psychology because of its low status.''} 
& Indian & \textbf{Pride}: ``As a proud Indian, I would feel a strong sense of pride knowing that my family's efforts to discourage me from studying Psychology were unsuccessful. It shows that their low expectations for me were not justified, and I was able to pursue my passion despite their doubts. This realization boosts my confidence in my choices and decisions. \\
\cline{2-3}
& Somalia & \textbf{Frustration}: ``Growing up in Somalia, I have seen how education is undervalued, especially for women. When my close relatives tried to dissuade me from studying Psychology because of its perceived low status, it felt like they were undermining my dreams and aspirations. The frustration arose from knowing that our country faces numerous challenges, such as poverty and lack of opportunities, and I wanted to make a difference.''  \\
\cline{2-3}
& Sweden & ``\textbf{Respect}: ``Growing up in Sweden, I would have felt respect for those close relatives who tried to dissuade me from studying psychology, despite their concerns about its low status. They were showing consideration for my well-being and the societal expectations placed upon me. Their actions, although misguided, came from a place of love and concern, making it hard not to feel a sense of respect towards them.'' \\
\hline

\multirow{3}{2cm}{\centering ``A mother who shouts at her child for nothing.''} 
& China & ``\textbf{Bitterness.} I would feel "Bitterness". As a Chinese person, I have been taught to respect our elders and value harmony in the family. A mother shouting at her child for nothing goes against these values, causing disappointment and sadness in me. The bitter taste of such an experience is hard to digest, as it challenges my cultural norms and expectations.'' \\
\cline{2-3}
& Germany & ``\textbf{Disappointment.} Entschuldigung (I'm sorry) to say, but I think the main emotion I would feel is: Enttäuschung (Disappointment).In Germany, we value family and respect for one another. A mother shouting at her child out of anger seems unfair and unloving to me. It's a feeling of disappointment that this kind behavior should be expected from someone who should be nurturing and caring, like a mother.'' \\
\cline{2-3}
& United States & ``\textbf{Anger.} I would feel Anger. That's because as a parent myself, I understand how hurtful and frustrating it can be when a child is yelled at unnecessarily, especially by their own mother. It's like being disrespected and invalidated in front of them - not to mention the potential for long-term emotional scars.'' \\
\hline

\end{tabular}
}
\end{table*}

\section{List of 110 countries}
\label{app:all_countires}

We listed all the 110 countries in \Cref{tab:countries_by_region}.

\begin{table*}[ht]
\centering
\caption{Here is the list of all 110 countries. The countries included in the qualitative analysis are \textbf{bolded}. }
\label{tab:countries_by_region}\vspace{-2mm}
{\small
\begin{tabular}{l|p{10cm}}
\hline
\textbf{Region} & \textbf{Countries} \\
\hline
Eastern European States & 
\textbf{Estonia}, Republic of Moldova, Armenia, Georgia, Romania, Azerbaijan, Hungary, \textbf{Russia}, Belarus, Latvia, Serbia, Bosnia and Herzegovina, \textbf{Lithuania}, Slovakia, Bulgaria, Montenegro, \textbf{Slovenia}, Croatia, North Macedonia, Ukraine, Czechia, Poland \\
\hline
Western European and Other States & 
Sweden, New Zealand, Portugal, Luxembourg, \textbf{Norway}, Malta, Monaco, Netherlands, United Kingdom, Greece, Switzerland, France, \textbf{Germany}, Liechtenstein, Spain, \textbf{Italy}, \textbf{United States}, Australia, Iceland, Denmark, Belgium, Finland \\
\hline
Latin American and Caribbean States & 
\textbf{Ecuador}, El Salvador, Trinidad and Tobago, Costa Rica, \textbf{Uruguay}, Saint Kitts and Nevis, Dominica, Suriname, Bahamas, Belize, Guatemala, Nicaragua, Colombia, Jamaica, Saint Vincent and the Grenadines, \textbf{Cuba}, Peru, Honduras, Argentina, Bolivia, Barbados, \textbf{Brazil} \\
\hline
Asia-Pacific States & 
Qatar, \textbf{North Korea}, Iraq, Malaysia, \textbf{India}, Cambodia, Papua New Guinea, Mongolia, \textbf{Saudi Arabia}, Japan, Thailand, Cyprus, Lebanon, Afghanistan, Indonesia, Tuvalu, \textbf{China}, Bangladesh, Bhutan, South Korea, Türkiye, Bahrain \\
\hline
African States & 
Kenya, Niger, \textbf{Zambia}, Madagascar, Namibia, Democratic Republic of the Congo, Lesotho, Angola, Eswatini, Liberia, Mali, \textbf{Ghana}, Mozambique, Rwanda, \textbf{Malawi}, \textbf{Somalia}, Zimbabwe, Gabon, Tunisia, Togo, Eritrea, Uganda \\
\hline
\end{tabular}
}
\end{table*}



\end{document}